\documentclass[10pt]{article} 

\usepackage[preprint]{tmlr}

\usepackage{amsmath}
\usepackage{amssymb}
\usepackage{amsthm}

\usepackage{algorithm}
\usepackage{algpseudocode}

\usepackage[usenames,dvipsnames]{xcolor} 
\usepackage{graphicx}
\usepackage{tikz}
\usepackage{svg}
\usepackage{subcaption}
\usepackage{wrapfig}
\usepackage{float}
\usepackage{placeins}
\usepackage{import}
\usepackage{booktabs}
\usepackage{multirow}
\usepackage{array}
\usepackage{threeparttable}

\usepackage{csquotes}
\usepackage{setspace}
\usepackage{xspace}
\usepackage{url}

\usepackage[activate={true},final,tracking=true]{microtype}

\usepackage{hyperref}
\hypersetup{
    colorlinks=true,
    linkcolor=NavyBlue,     
    citecolor=NavyBlue,     
    urlcolor=NavyBlue,      
    breaklinks=true,        
    pdfpagemode=UseNone,    
    pdfstartview=FitH,      
    pdftitle={CODE: A global approach to ODE dynamics learning}, 
    pdfauthor={Wildt et al.}, 
    pdfsubject={Machine Learning, Dynamical Systems},
    pdfkeywords={ODE, Dynamics Learning, Scientific Machine Learning}
}

\usepackage[capitalise]{cleveref}

\usepackage{amsmath,amsfonts,bm}

\def\eqref#1{equation~\ref{#1}}

\def\1{\bm{1}}

\DeclareMathAlphabet{\mathsfit}{\encodingdefault}{\sfdefault}{m}{sl}
\SetMathAlphabet{\mathsfit}{bold}{\encodingdefault}{\sfdefault}{bx}{n}

\definecolor{anth}{RGB}{62,68,76}
\definecolor{blue}{rgb}{0,0.5,1}
\definecolor{brown}{rgb}{0.2,0,0}
\definecolor{commentBlue}{rgb}{0.0,0.0,0.8}
\definecolor{cyan}{rgb}{0.2,0.8,0.8}
\definecolor{darkblue}{rgb}{0.0,0.0,0.3}
\definecolor{green}{rgb}{0,0.6,0.3}
\definecolor{gruen}{RGB}{0,150,0}
\definecolor{lblue}{RGB}{0,190,255}
\definecolor{lila}{rgb}{0.49400,0.18400,0.55600}
\definecolor{mblue}{RGB}{0,81,158}
\definecolor{orange}{rgb}{1,0.5,0}
\definecolor{red}{rgb}{1,0.2,0.2}
\definecolor{rot}{rgb}{1,0.12500,0.009800}

\newcommand{\code}{C\hspace*{0.07em}ODE\xspace}

\newcommand{\diff}{\,\mathrm{d}} 
\newcommand{\eg}{\textit{e}.\textit{g}.\xspace} 

\newcommand{\ie}{\textit{i}.\textit{e}.\xspace} 

\newcommand{\representer}{\ensuremath{\tilde{f}}}

\newcommand*{\mat}[1]{{\boldsymbol{\mathbf{#1}}}}
\newcommand*\dd{\mathop{}\!\mathrm{d}}

\newcommand{\ra}[1]{\renewcommand{\arraystretch}{#1}}

\pdfpageattr{/Group <</S /Transparency /I true /CS /DeviceRGB>>}

\newlength{\alphabet}
\title{\code: A global approach to ODE dynamics learning}

\def\month{MM}
\def\year{YYYY}

\author{%
    \name Nils Wildt \email nils.wildt@iws.uni-stuttgart.de \\
    \addr Institute for Modelling Hydraulic and Environmental Systems\\
    University of Stuttgart
    \AND
    \name Daniel M. Tartakovsky \email tartakovsky@stanford.edu \\
    \addr Department of Energy Science and Engineering\\
    Stanford University
    \AND
    \name Sergey Oladyshkin \email sergey.oladyshkin@uni-stuttgart.de \\
    \addr Institute for Modelling Hydraulic and Environmental Systems\\
    University of Stuttgart
    \AND
    \name Wolfgang Nowak \email wolfgang.nowak@iws.uni-stuttgart.de \\
    \addr Institute for Modelling Hydraulic and Environmental Systems\\
    Cluster of Excellence SimTech \\
    University of Stuttgart
}

\begin{document}

\maketitle

\begin{abstract}
    Ordinary differential equations (ODEs) are a conventional way to describe the observed dynamics of physical systems. Scientists typically hypothesize about dynamical behavior, propose a mathematical model, and compare its predictions to data. However, modern computing and algorithmic advances now enable purely data-driven learning of governing dynamics directly from observations.\smallskip

In data-driven settings, one learns the ODE's right-hand side (RHS). Dense measurements are often assumed, yet high temporal resolution is typically both cumbersome and expensive. Consequently, one usually has only sparsely sampled data. In this work we introduce ChaosODE (\code), a Polynomial Chaos ODE Expansion in which we use an arbitrary Polynomial Chaos Expansion (aPCE) for the ODE's right-hand side, resulting in a global orthonormal polynomial representation of dynamics. We evaluate the performance of \code in several experiments on the Lotka-Volterra system, across varying noise levels, initial conditions, and predictions far into the future, even on previously unseen initial conditions. \code exhibits remarkable extrapolation capabilities even when evaluated under novel initial conditions and shows advantages compared to well-examined methods using neural networks (NeuralODE) or kernel approximators (KernelODE) as the RHS representer. We observe that the high flexibility of NeuralODE and KernelODE degrades extrapolation capabilities under scarce data and measurement noise.\smallskip 

Finally, we provide practical guidelines for robust optimization of dynamics-learning problems and illustrate them in the accompanying code.
\end{abstract}

\section{Introduction}\label{sec1}
Scientists strive to understand the mechanisms of the world. In mathematical modeling, mechanisms are often formulated as ordinary differential equations (ODE) dynamics. Let us assume a model \(\frac{\mathrm{d}}{\mathrm{d}t} x(t) = f(x(t)).\) The state \(x(t) \in \mathbb{R}^n\) evolves in time \(t\in [t_0,t_{end}]\) according to a continuous dynamic \(f: \mathbb{R}^n \mapsto \mathbb{R}^n.\) Very often, the concrete dynamic \(f(x(t),\theta)\) also depends on several parameters \(\theta\) that alter the dynamical behavior. To calibrate the model, numerous techniques have been developed in recent years to determine the function that best fits the observed data. Typically, scientists propose models that depend on various parameters. In the simplest case, the parameters have physical meanings and can be determined through measurements. If the parameters are either not observable by measurements or have no physical measurable meaning, they need to be found by inversion and optimization techniques. The model is then fitted on the basis of a performance metric \eg minimizing the pointwise distance between the model output and real-world observations \citep[\eg][]{tarantola2005inverse,vandeschootBayesianStatisticsModelling2021}. If the model is not suitable, one iterates model selection and improvement \citep{hoge2018primer}. \smallskip

Relaxing the assumption of a fixed parametric model, we may search over a function space for the RHS. In doing so, we can use optimization to discover the dynamics in a very general way. However, we assume that we only have state measurements at discrete points in time and no inherent derivative information of the states themselves. Hence we cannot just solve a simple regression problem for the RHS, but need to optimize in a highly nonlinear setting in which each evaluation of the model involves the time integration of the current estimated dynamics. The choice of the RHS function space (ansatz) governs the approximation power and the training complexity. In the work on NeuralODE, modeling of arbitrary learnable operators was widely seen for the first time \citep{chenNeuralOrdinaryDifferential2019a}. This idea was later formally generalized for arbitrary right-hand sites and termed Universal Differential Equations (UDE) by \citet{rackauckas2020universal}. Within this general UDE setting, the RHS representation is itself a key modeling decision. In succession to the idea of NeuralODEs, over the past few years, a thorough comprehension of the dynamics involved in learning Ordinary Differential Equations (ODEs) via adequate datasets has been developed. This enhanced understanding has increasingly shown its practical applications in a multitude of fields, particularly within geosciences \citep{Shen_Appling_Gentine_Bandai_Gupta_Tartakovsky_Baity-Jesi_Fenicia_Kifer_Li_et_al_2023} and hydrology \citep{Höge_Scheidegger_Baity-Jesi_Albert_Fenicia_2022}. In parallel, other RHS representations were investigated, specifically the use of Gaussian Processes, which are another known nonlinear approximation method in machine learning applications \citep{kanagawa2018gaussianprocesseskernelmethods,hegde2022variational}. As these approaches only rely on the mean path, we refer to these methods with KernelODE, highlighting the underlying theory of kernel methods \citep{Santin2021}. Both neural network- and kernel-based frameworks typically require dense training datasets containing hundreds of data points. However, many practical applications, such as contaminant transport of TCE \citep{wust1999combined} and PFAS \citep{croll2022adaptation}, commonly have far fewer observations. In such sparse settings, highly flexible models risk poor extrapolation without strong inductive bias.
\smallskip

While NeuralODEs have gained widespread use in scientific applications, the simplicity often portrayed in studies does not align with the difficult challenges practitioners encounter when trying to effectively optimize dynamics using limited or noisy data. A notable issue is that tests on out-of-distribution, previously unseen initial states are rarely conducted, leaving the ability to extrapolate largely unexplored. 

We address this challenge of learning from limited data by introducing arbitrary Polynomial Chaos ODE Expansion (\code), a new global and orthonormal polynomial RHS approximation. This approach naturally represents polynomial dynamics (common across physics, chemistry, and biology) and can approximate arbitrary functions. In \code, we find a multinomial expansion to represent the dynamics of differential equations. This is realized using the arbitrary Polynomial Chaos (aPC) Expansion as described in \citep{oladyshkin2012data}, an approach built on the generalized Polynomial Chaos expansion \citep{xiu2002wiener}. 
In contrast to a full polynomial basis \citep{fronkInterpretablePolynomialNeural2023}, our data-driven basis is orthogonal with respect to the data distribution. In our case, this is the distribution of the states that we know over time. The ChaosODE results in relatively few coefficients that determine the behavior of the RHS while enabling the calculation of dynamics across the entire allowable domain, since polynomials are global functions. Previous work on improving UDEs in time-only settings projected the NeuralODE solution onto an orthogonal basis \citep{de2023anamnesic}, while PCE collocation learning was applied for physics infusion by \citep{sharma2024physics}, although their approach focused on general (collocation) PDE solving and surrogate modeling rather than time-only applications.\smallbreak

Overall, we hypothesize that problem-tailored RHS spaces improve learning. Therefore, we aim to answer the following unresolved questions:

\begin{enumerate}
\item How do current UDE learning techniques perform when training data is both scarce and noisy?
\item What extrapolation and generalization capabilities do KernelODE and NeuralODE exhibit, and how does \code's global approximation structure compare in these aspects?
\item Does the orthonormal structure of the dynamics representation facilitate learning in highly non-linear systems?
\end{enumerate}

We first present \code, followed by the other methods already established in \Cref{sec:methods}. Then we illustrate the training procedure followed by the results and a discussion (\Cref{sec:results}) to be concluded in \Cref{sec:conclusion}.

\FloatBarrier

\section{Methodology}
\label{sec:methods}
This section provides a complete explanation of the general RHS learning problem, introduces the \code framework and comparison methods, and provides training and implementation specifics.
\subsection{From data to an ODE}
Consider a system whose states we observe over time. The observed data are denoted as \( \mathcal{D} = \{(t_i, \hat{\mat{x}}_i)\} \) for \( i = 1, \ldots, N \), where \( N \) represents the number of distinct measurements taken at arbitrary times and sorted in ascending order. Measurements typically contain noise \( \varepsilon \), such that \( \hat{\mat{x}}_i = \mat{x}_i + \varepsilon_i \), where \( \mat{x}_i = \mat{x}(t_i) \) represents the true unknown value without noise.

Since the time intervals \( \Delta t = t_i - t_{i-1} \) can be arbitrary, we cannot directly approximate the derivative using
\(
\frac{\dd}{\dd t} x_i \approx \lim_{\Delta t \rightarrow 0} \frac{x_i - x_{i-1}}{\Delta t}.
\)
Instead, we assume that the state evolution follows a differential equation. Specifically, we consider an autonomous ordinary differential equation (ODE) of the form
\begin{equation}
\frac{\dd}{\dd t} \mat{x}(t) = \dot{\mat{x}}(t) = f(\mat{x}(t)),
\end{equation}
where \( \mat{x}(t) \in \mathbb{R}^n \) denotes the state of the system at time \( t \in \mathbb{R} \) and \( f(\mat{x}(t)) \) is the vector field. When combined with an initial state \( (t_0, \mat{x}_0) \), this formulation defines an initial value problem (IVP) \citep{bronstein1989taschenbuch}. 

Such IVPs are solved with \begin{equation}
  \mat{x}(t) = \int\limits_{t_0}^{t} f(\mat{x}(\tau)) \dd \tau + \mat{x}_0.
  \label{eq:s1_integration}
\end{equation} The ODE is determined by the dynamics \(f(\cdot)\) and an initial condition. Ultimately, the goal is therefore to find an approximation of the flow field \(\representer_{{\theta}\prime}(\cdot)\approx f(\cdot)\) and an estimation of \(\mat{x}(t_0)\approx\Tilde{\mat{x}}(t_0)=\Tilde{\mat{x}}_0.\) We assume that \(\representer(\cdot)\) is parameterized by \(\theta\prime\) and, in general, we can determine the dynamics and initial conditions by finding \[\theta = \begin{pmatrix}
  \theta\prime \\ \mat{x}_0
\end{pmatrix},\] in the general case of unknown \(x_0.\) In the following, we call \(\representer_{\theta\prime}(\cdot)\) the representation of the RHS. \smallskip
The optimal parameters are obtained by solving the optimization problem:
\begin{align}
\theta^* = \arg\min_{\theta} \ell_d,
\label{eq:parameter_optimization}
\end{align}
where $\tilde{\mathbf{x}}_i(\theta)$ denotes the prediction of the model at the measurement point $i$ as a function of the parameters $\theta$ and the distance loss between the observed data at several times and the prediction at the same times, for example, $\ell_d=\sum\limits_{i=1}^N \|{\mat{x}_i - \tilde{\mat{x}}_i}\|_d$. 
Next, we elaborate on different approaches to model $\tilde{f}(\cdot).$

\paragraph{Choice of a basis for the vector field}

The choice of representation space fundamentally determines the quality and generalization of the approximation. To approximate functions with specific properties, the hypothesis class must be able to express those properties.
Consider the practical implication: if we seek to approximate physically valid right-hand sides (RHS) of differential equations, then the chosen representation space must be capable of encoding physical constraints and behaviors. The representation space acts as a constraint on the solution space and it cannot approximate what the chosen framework cannot express.
This principle motivates our comparative analysis of three distinct RHS representation approaches.

\begin{itemize}
  \item NeuralODE (see \cref{ssec:neuralode}): Neural networks encode the dynamics, leveraging universal approximation capabilities. 
  \item KernelODE (see \cref{ssec:kernelode}): Gaussian process mean trajectories represent dynamics, provide smoothness guarantees, and localized basis function.
  \item ChaosODE (see \cref{ssec:chaosode}): Polynomial basis representation offers analytical tractability and global approximation.
\end{itemize}
Each representation space imposes different structural assumptions and constraints, directly influencing both the quality and characteristics of the resulting approximation. Thus, the representation is a fundamental modeling decision that determines the space of achievable solutions, not merely a technical implementation detail.

Next, we summarize the existing approaches and introduce a novel ChaosODE approach.

\subsection{NeuralODE}\label{ssec:neuralode}
Neural Ordinary Differential Equations (NeuralODE) leverage neural networks to learn dynamic representations \citep{chenNeuralOrdinaryDifferential2019a}. Building on the foundational work in neural networks \citep{hopfield1982neural,Rumelhart_Hinton_Williams_1986}, the core concept is elegantly simple: a neural network serves as the regressor for the RHS side of an ODE. 

Formally, the learned function is defined as \(\representer: \mathbb{R}^n \mapsto \mathbb{R}^n\), where \(n\) denotes the number of states. The NeuralODE framework has inspired researchers in diverse disciplines to integrate ODEs with artificial neural networks as universal function approximators \citep{rackauckas2020universal}. 

The selected network architecture should capture the essential features of the expected dynamics. According to the universal approximation theorem, it is theoretically possible to model arbitrary dynamics using this approach \citep{Hornik_1991}. However, the choice of network architecture and hyperparameters significantly influences the performance of the model. The many small choices that finally determine the dynamics in a NeuralODE setting often require timely fine-tuning and can easily induce researchers' bias towards certain dynamical behavior.

In general, NeuralODEs offer remarkable flexibility in modeling dynamical systems. Although this approach can arbitrarily finely interpolate within the training data domain, extrapolation remains arbitrary, as we lack understanding of the effective ansatz function generated by the network combination that represents the dynamics. This expansive solution space, while powerful for interpolation tasks, requires careful consideration when predicting system behavior beyond the observed regime. Consequently, extrapolation beyond the training data can be unreliable, as neural-network parameterizations often generalize poorly outside the data support.

\subsection{Kernel ODE}\label{ssec:kernelode}
The second approach employs kernel methods to approximate the RHS of an ODE. Like neural networks, kernels provide a regression framework for learning dynamics from data. The key distinction lies in their construction: Kernel methods build the approximation using basis functions centered at selected data points,which makes the method inherently local.

A \textit{kernel} is a function \( k: \mathcal{X} \times \mathcal{X} \rightarrow \mathbb{R} \) that measures the similarity between inputs. For all \( x, x' \in \mathcal{X} \), the kernel satisfies:
\begin{equation}
k(x, x') = \langle \phi(x), \phi(x') \rangle_{\mathcal{H}}
\end{equation}
where \(\phi:\mathbb{R}^d \mapsto \mathcal{H} \) maps the input to a Hilbert space \(\mathcal{H}\) \citep{hofmann2008kernel,Santin2021,schaback2006kernel}. This mapping enables nonlinear regression without explicitly computing high-dimensional features.

The dynamics approximation uses a kernel expansion:
\begin{equation}
\representer^{\theta^{\mathcal{K}}}_{1,\mathcal{K}}(x) = \sum_{i=1}^n c^1_i k(x, x_i)
\end{equation}
Here, \(c^1_i \in \mathbb{R}\) are coefficients and \(\{x_i\}_{i=1}^n\) are collocation points (also called inducing or pilot points \citep{schaback2006kernel,titsias09a,doherty2011approaches}). These collocation points define where the kernel basis functions are centered. The approximation quality depends directly on their number and placement, as the method interpolates between these points using the kernel-weighted combination.

In the ideal case where we know the true dynamics \(f\) at collocation points, the target vector is \(\theta^{\mathcal{K}} = [\theta_1, \theta_2, \ldots, \theta_n]^\top \) with \(\theta_i = f(x_i)\). The coefficients are then computed through:
\begin{equation}
c = (K + \lambda I)^{-1} \theta^{\mathcal{K}}
\end{equation}
The kernel matrix \(K\) has entries \( K_{ij} = k(x_i, x_j) \) and is positive definite by construction. The regularization parameter \( \lambda > 0 \) prevents overfitting and ensures numerical stability, with \( I \) denoting the identity matrix.

In practice, we do not have access to the true dynamics at collocation points. Therefore, the unknown collocation values \(\theta^{\mathcal{K}}\) are considered to be trainable parameters. During training, these collocation values \(\theta^{\mathcal{K}}\) should be adjusted so that the resulting ODE solution matches the observed data. Given \(\theta^{\mathcal{K}}\), coefficients follow directly from the linear system \(c = (K + \lambda I)^{-1} \theta^{\mathcal{K}}\). The optimization thus focuses on finding collocation values that produce dynamics consistent with measurements, while the kernel framework handles interpolation between collocation points.

For systems with multiple state variables (\(n > 1\)), we construct separate kernel regressors for each dimension. Each regressor maps \(\mathbb{R}^n \mapsto \mathbb{R}\). For a two-dimensional system (\(n=2\)), the complete approximation becomes:
\begin{equation}
  \representer^{\theta^{\mathcal{K}}}_{\mathcal{K}}(x) := \begin{pmatrix}
    \representer^{\theta^1}_{1,\mathcal{K}}(x) \\
    \representer^{\theta^2}_{2,\mathcal{K}}(x)
  \end{pmatrix}
\end{equation}

The data-dependent nature of kernel methods affects the approximation quality across the state space. Near collocation points, the approximation typically achieves high accuracy. Between collocation points, the kernel provides smooth interpolation. Outside the region covered by pilot points, the extrapolation behavior depends entirely on the chosen kernel function. Kernel-specific hyperparameters (such as length scales) and the regularization parameter \(\lambda\) require tuning through cross-validation or similar methods.

While some implementations frame this approach within a Gaussian Process (GP) framework \citep{heinonen2018learning}, the deterministic kernel regression presented here is equivalent to using the mean prediction of a GP \citep{kanagawa2018gaussianprocesseskernelmethods}. Probabilistic interpretation adds uncertainty quantification, but does not change the fundamental regression mechanism. As a local method, extrapolation behavior is governed by the kernel choice and its length scale(s), which then corresponds to a notion of smoothness in the dynamic space.

\subsection{Chaos ODE}\label{ssec:chaosode}
In this work, we introduce Chaos ODE that employs data-driven orthononmal polynomial chaos basis functions \citep{oladyshkin2012data} to approximate the RHS of an ODE. Unlike its traditional role in uncertainty quantification, we re-purpose aPC as a data-driven global and orthonormal basis over the state space for learning dynamics.

The dynamics \(f(\cdot)\) are approximated through an expansion in orthogonal polynomials:
\begin{equation}
  f(x) = \sum\limits_{p=0}^{\infty} c_p \Phi_p(x),
\end{equation}
where \(x \in \mathbb{R}^n\) denotes the state variable and \(\{\Phi_p\}_{p \in \mathbb{N}_0}\) forms an orthonormal polynomial basis. These basis functions satisfy the orthogonality condition with respect to a weight function \(w(x)\) induced by a probability measure \(\mathcal{P}\):
\begin{equation}\label{eq:inner_prod_ortho}
  \left\langle\Phi_p, \Phi_q\right\rangle_{\mathcal{P}}:=\int_{\mathbb{R}^n} \Phi_p(x) \Phi_q(x) \, w(x) \, \mathrm{d}x = \int_{\mathbb{R}^n} \Phi_p(x) \Phi_q(x) \, \mathrm{d}\mathcal{P}(x) = \delta_{pq}, \quad \forall p, q \in \mathbb{N}_0,
\end{equation}
where \(\delta_{pq}\) denotes the Kronecker delta. The expansion coefficients are obtained via projection:
\begin{equation}
  c_p = \langle f, \Phi_p\rangle_{\mathcal{P}}, \quad p \in \mathbb{N}_0.
\end{equation}

The probability measure \(\mathcal{P}\) defines the weight function for orthonormality. In classical Wiener polynomial chaos expansion, the orthogonal family is fixed by the input distribution, \ie Hermite polynomials for Gaussian distribution. Generalized polynomial chaos extends this mapping across the Askey scheme to cover additional standard distributions (\eg Legendre for uniform, etc.: \citep{xiu2002wiener}). In contrast, we employ arbitrary polynomial chaos \citep{oladyshkin2012data}, constructing data-driven orthonormal polynomials from empirical moments of \(\mathcal{P}\). For practical computation, we truncate the expansion at polynomial degree \(N_{\text{max}}\):
\begin{equation}
  \representer^{\theta^{\mathcal{C}}}_{\mathcal{C}}(x) \approx \sum\limits_{p=0}^{N_{\text{max}}} \theta^{\mathcal{C}}_p \Phi_p(x),
\end{equation}
where \(\theta^{\mathcal{C}}_p\) are the parameters to be learned. The truncation degree \(N_{\text{max}}\) balances the approximation accuracy against the computational cost. For \(n\)-dimensional systems, multivariate polynomials are constructed through tensor products, with the total number of basis functions growing combinatorially with dimension and polynomial degree.

This approach provides a global approximation with spectral convergence rates for smooth dynamics. Unlike local methods such as kernels, polynomial chaos expansions use global basis functions spanning the entire state space. 
Although there are standard polynomial families for common distributions \citep{sullivan2015introduction}, we construct data-driven orthonormal polynomials using arbitrarily chaotic polynomial chaos. This approach adapts the basis to the empirical distribution of the state variables.

We construct each polynomial \(\Phi_d\) of degree \(d\) as:
\begin{equation}
  \Phi_d(x)=\frac{1}{\sqrt{h_d}} \sum_{i=0}^d c_i x^i,
\end{equation}
where \(h_d\) is a normalization constant and \(c_0, \ldots, c_d\) are coefficients to be determined. To find these coefficients, we enforce orthogonality through a moment-matching approach \citep{oladyshkin2012data}.

The orthogonality constraint \eqref{eq:inner_prod_ortho} requires that \(\langle \Phi_i, \Phi_j \rangle_{\mathcal{P}} = \delta_{ij}\). Combined with the auxiliary condition \(c_d = 1\) (fixing the leading coefficient), this yields a linear system. Using the \(k\)-th raw moment \(\mu_k = \int_{\mathbb{R}}x^k \, \mathrm{d}\mathcal{P}(x)\), we obtain:
\begin{align} \label{eq:lin_sys_moments}
\begin{bmatrix}
\mu_0 & \mu_1 & \cdots & \mu_{d-1} \\
\mu_1 & \mu_2 & \cdots & \mu_{d} \\
\vdots & \vdots & \ddots & \vdots \\
\mu_{d-1} & \mu_d & \cdots & \mu_{2d-2} \\
0 & 0 & \cdots & 1
\end{bmatrix}
\begin{bmatrix}
c_0 \\
c_1 \\
\vdots \\
c_{d-1} \\
c_d
\end{bmatrix}
=
\begin{bmatrix}
0 \\
0 \\
\vdots \\
0\\
1
\end{bmatrix}.
\end{align}

This Hankel matrix system determines the coefficients for the \(d\)-th polynomial using moments of up to order \(2d-1\). When the distribution \(\mathcal{P}\) is unknown, we estimate the moments from the data:
\begin{equation}
  \tilde{\mu}_k = \frac{1}{N} \sum_{j=1}^{N} x_j^k,
\end{equation}
where \(\{x_j\}_{j=1}^N\) are the observed state values. By the law of large numbers, \(\tilde{\mu}_k \to \mu_k\) as \(N \to \infty\).

To ensure numerical stability and maintain orthonormality, we normalize each polynomial after construction:
\begin{equation}
  \Phi_d(x) = \frac{\Phi_d(x)}{\|\Phi_d\|_{L^2(\mathcal{P})}},
\end{equation}
where the norm is estimated using Monte Carlo integration:
\begin{equation}
\|\Phi_d\|_{L^2(\mathcal{P})}^2 \approx \frac{1}{N} \sum_{j=1}^{N} [\Phi_d(x_j)]^2.
\end{equation}
This normalization prevents numerical instabilities that arise when polynomials of increasing degree have vastly different magnitudes.

For multivariate systems with \(\mathbf{x} \in \mathbb{R}^n\), we construct tensor product polynomials:
\begin{equation}
\Phi_{\boldsymbol{\alpha}}(\mathbf{x}) = \prod_{i=1}^n \Phi_{\alpha_i}(x_i),
\end{equation}
where \(\boldsymbol{\alpha} = (\alpha_1, \ldots, \alpha_n)\) is a multi-index with total degree \(|\boldsymbol{\alpha}| = \sum_{i=1}^n \alpha_i\). Restricting to polynomials with \(|\boldsymbol{\alpha}| \leq N_{\text{max}}\) yields
\begin{equation}
M = \binom{n + N_{\text{max}}}{n} = \frac{(n + N_{\text{max}})!}{n! N_{\text{max}}!}
\end{equation}
basis functions. This combinatorial growth with dimension \(n\) limits the method to moderate-dimensional problems.

\par

Overall we now have three different RHS approximators that can be trained, with an overview in \cref{tab:overview_RHS}
\begin{table*}[htbp]
  \centering
  \caption{Overview of different RHS approximation approaches.}
  \label{tab:overview_RHS}
  \begin{tabular}{p{2.0cm} p{4cm} p{3.5cm} p{3.5cm} }
    \toprule
    \textbf{Name} & \textbf{Representation} & \textbf{Trainable parameters} & \textbf{Hyperparameters} \\
    \midrule
    NeuralODE & \(\representer^{\theta^{\mathcal{N}}}_{\mathcal{N}}(x) := \mathcal{NN}^{\theta^{\mathcal{N}}}(x)\) & Weights and biases & Network architecture, activation functions \\
    \addlinespace[1.0em]
    ChaosODE & \(\representer^{\theta^{\mathcal{C}}}_{\mathcal{C}}(x) :=\begin{pmatrix}
    \boldsymbol{\varphi}_1^N(x) \theta_1^{\mathcal{C}} \\
    \boldsymbol{\varphi}_2^N(x) \theta_2^{\mathcal{C}}
    \end{pmatrix}\)  & Expansion coefficients & Maximal polynomial degree \\ 
    \addlinespace[1.0em]
    KernelODE & \(\representer^{\theta^{\mathcal{K}}}_{\mathcal{K}}(x) := \begin{pmatrix}
    \representer^{\theta^1}_{1,\mathcal{K}}(x) \\
    \representer^{\theta^2}_{2,\mathcal{K}}(x)
    \end{pmatrix}\) & Pilot values at pilot points & Number \& location of pilot points, Kernel function\\
    \bottomrule
  \end{tabular} 
\end{table*}

\subsection{Training technicalities and guidelines}
Training high-dimensional surrogates with neural networks is well-established. Gradient-based optimizers such as Adam \citep{diederik2014adam} with properly tuned learning rates consistently find good optima. Beyond achieving low training losses, these models must also generalize to unseen data. 

In the ODE context, generalization takes two distinct forms: predicting state estimates for unseen initial conditions ($\mat{x}_0$) and for unseen time points. The first requires an ansatz space that correctly extrapolates functional behavior beyond the training domain. The second (temporal) type relies on accurate integration over longer horizons once the RHS is learned. In all three setups, the RHS can theoretically be easily determined if we could evaluate $\mat{f}(x(t))$ at any time. However, as we assume there are only a few observations, an optimization determines the interpolation and extrapolation. 
Optimizing through ODE solutions presents three key challenges in our setting:

\begin{enumerate}
    \item {Parameter sensitivity:} Small parameter changes can produce large trajectory variations. This sensitivity requires the optimizer to accept poor intermediate solutions while traversing unfavorable regions of the loss landscape.
    
    \item {Solver divergence:} When the ODE solver diverges, it produces neither solutions nor gradients, preventing informed parameter updates.
    
    \item {RHS-dependent dynamics:} Different right-hand side formulations create distinct parameter-trajectory dependencies, complicating the development of a unified optimization strategy.
\end{enumerate}

In the following paragraphs, we describe how we obtain stable training for all three RHS choices. To make the three approaches comparable, we succeeded in identifying one optimization approach for all scenarios.

\paragraph{Multiple- vs. single shooting}
Single-shooting methods solve the entire ODE system from $x_0$ over the full time interval $t \in [t_0, t_{end}]$ using a single set of parameter coefficients. This approach faces a fundamental stability bottleneck: if the current parameter estimates of the RHS cause numerical instabilities (non-finite values) at any point during integration, the entire solution trajectory becomes invalid, preventing objective function evaluation and halting optimization progress. Although present for all three RHS choices, this is especially severe in the \code setting, as polynomials grow unboundedly outside their stable regions, rapidly producing overflow errors that terminate integration.
\smallskip

This creates a circular dependency problem. Successful parameter updates during training require stable numerical integration to compute gradients and objective values. However, achieving numerical stability often requires parameter values that are already close to the optimal solution. Without good initial parameter guesses, the optimization becomes trapped in overly conservative parameter regions that produce stable but overly smooth, mean-like solutions instead of capturing the dynamic behavior present in the observed data. 
\smallskip

Although global optimization methods can potentially escape these conservative regions, they face the same stability constraints. Even when exploration successfully identifies parameter regions that yield stable trajectories, subsequent exploitation phases must navigate carefully to avoid reentering unstable regimes. In practice, this makes it extremely difficult for single-shooting approaches to converge to solutions that accurately capture non-trivial dynamics without requiring exceptional initial parameter estimates.
\smallskip

Multiple-shooting methods for ODEs \citep{turan2021multiple} prevent solver divergence by partitioning the integration interval into smaller subintervals, avoiding non-finite values during integration. Although multiple shooting requires careful initialization of each segment’s initial values and enforcement of continuity constraints, we employ a hybrid approach that bypasses this complexity. Multiple shooting generates initial approximations that, though segmented and non-smooth, provide robust starting points. A subsequent single-shooting refinement step produces smooth and accurate solutions through gradient-based optimization. This two-stage strategy combines numerical stability with high fit accuracy, as our tests confirm. Sequential optimization approaches, which increase the number of data segments, consistently underperformed compared to this direct multiple-shooting method in our preliminary tests.

\paragraph{Discretize then Optimize vs. Optimize then Discretize}
We adopt a discretize-then-optimize approach \citep{onken2020discretize, sapienza2024differentiable} for robust optimization. This method discretizes the ODE using a fourth-order Runge-Kutta (RK4) scheme, then applies automatic differentiation through the discretized operations. This contrasts with optimize-then-discretize (adjoint) methods, which derive continuous gradients before numerical discretization. Our choice follows recent findings that discretize-then-optimize provides comparable accuracy, along with a simpler implementation and faster computation for our problem class, which our experiments confirm.

\paragraph{Initial RHS Parameter Estimation}
Despite the application of global optimization techniques, accurate initial parameter estimates for the respective RHS representative remain crucial for convergence efficiency. We exploit the regularity typically present in time-series data by approximating the solution with a kernel regression surrogate. Since kernel regression is closed under linear operations, we can compute time derivatives at arbitrary points throughout the dataset. This provides a known surrogate $\tilde{f}_{\text{kernel}}(\cdot)$ that can be directly evaluated, thereby transforming the problem into a simple regression scenario. Using these derivative estimates to conduct preliminary training of all RHS models, we obtain initial coefficients that avert immediate divergence during the subsequent optimization phase.

\paragraph{Optimization Pipeline}
From these initial parameter values, we perform a sequence of optimization steps, each balancing numerical stability, exploration, and exploitation. First, we use a Particle Swarm Optimization (PSO) algorithm \citep{eberhart1995particle} in a single-shot setting to move from arbitrary parameter initializations to a stable solution that does not diverge. This is possible, as the initial parameter estimation allows for plausible solutions. The PSO is well suited here, as initial infeasible parameter configurations will simply drop out, leaving us with those that produce stable solutions. In addition, it is fast, allowing for high exploration capabilities. From then on, we employed the Covariance Matrix Adaptation Evolution Strategy (CMA-ES) \citep{hansen2001completely} in a multiple shooting setup. CMA-ES is a gradient-free evolutionary strategy. It provides multistart optimization capabilities while adapting the search distribution toward more favorable solutions. This allows exploration within the feasible parameter space, minimizing the summed losses across all segments. It is again quite robust to intermediate infeasible solutions, as they would again just be dropped in favor of newly drawn coefficient combinations. These two methods conclude the numerically stable and exploratory parts. After CMA-ES converged, we used a Quasi-Newton method \citep{nocedal2006numerical} in the multiple-shooting setup. This exploits the local gradients as well as an approximation of the Hessian and, in most cases, converges to locally approximated parameter sets. Finally, we smooth the segmented multiple-shooting solution via a single-shooting Quasi-Newton refinement. All optimization was performed using the Manopt.jl and Optim.jl julia libraries \citep{Bergmann2022,Mogensen2018}. 

\FloatBarrier

\section{Results and discussion}
\label{sec:results}
Traditional model calibration optimizes a limited number of physical parameters to minimize data loss while assuming a fixed functional form. Recent advances in machine learning enable a more powerful approach; we can simultaneously learn the functional dependencies during optimization. We learn a plausible dynamic within a space of possible dynamical systems.  Neural Ordinary Differential Equations (NeuralODEs) have driven recent progress in this field. Their high flexibility enables them to approximate arbitrary dynamics. However, this flexibility comes at a cost: unclear extrapolation capabilities beyond the training domain.

We demonstrate how selecting a suitable basis can substantially improve  extrapolation to new initial conditions. To this end, we evaluate the recently  introduced \code and related methods on the Lotka–Volterra benchmark  \citep{lotka1925,volterra1926} for learning dynamics. We compare ChaosODE, NeuralODE, and KernelODE across the four scenarios S1–S4 described in \cref{tab:scenario_metrics}.

\begin{table*}[htbp]
  \centering
  \ra{0.95}
  \begin{threeparttable}
    \caption{Experimental scenarios for polynomial chaos expansion modeling}
    \label{tab:scenario_metrics}
    \begin{tabular}{p{1.5cm} p{4.0cm} p{7.5cm}}
    \toprule
    \textbf{Scenario ID} & \textbf{Scenario Type} & \textbf{Description} \\
    \midrule
    S1 & Perfect information baseline&
    Optimization from globally optimal initial coefficients on 144 training observations to evaluate how local optimization affects extrapolation capabilities without noise. \\
    \addlinespace[0.5em]
    S2 & Data effect & 
    Systematic increase in training data size to identify minimum data requirements for reliable model performance without noise.\\
    \addlinespace[0.5em]
    S3 & Noise effect & 
    Analyze the impact of data noise on the training robustness. \\
    \addlinespace[0.5em]
    S4 & ChaosODE advantage & 
    Investigate ChaosODE, test orthonormal- vs. standard- basis.\\
    \bottomrule
    \end{tabular}
  \end{threeparttable}
\end{table*}
\FloatBarrier

\subsection{Performance Metrics}
Given that the state magnitudes in our benchmark problem are of comparable scale, we employ the mean squared error (MSE) as our evaluation metric:
\begin{equation}
\text{MSE} = \frac{1}{n} \sum_{i=1}^{n} (x_i - \hat{x}_i)^2
  \label{eq:mse}
\end{equation}
where $x_i$ represents the simulation result, and $\hat{x}_i$ denotes the corresponding observed data.\par

\begin{table*}[hbtp]
  \centering
  \ra{0.95}
  \begin{threeparttable}
    \caption{Performance evaluation setup for dynamic system modeling}
    \label{tab:performance_metrics}
    \begin{tabular}{p{1.5cm} p{2.5cm} p{4cm} p{6cm}}
    \toprule
    \textbf{Comparison ID} & \textbf{Comparison Type} & \textbf{Description} & \textbf{Purpose} \\
    \midrule
   \textit{ex-it} & In training time &
    Evaluation using original training dataset &
    Measures in-distribution training performance \\
    \addlinespace[0.5em]
    \textit{ex-oot} & Out of training time & Testing with lengthened time intervals but same $x_0$ as in \textit{ex-it} &
    Assesses model's ability to extrapolate dynamics over longer durations \\
    \addlinespace[0.5em]
    \textit{ex-ood} & Out of distribution &
    Testing with new, different initial conditions for prolonged time &
    Evaluates model's generalization to unseen starting points \\
    \bottomrule
    \end{tabular}
  \end{threeparttable}
\end{table*}
\FloatBarrier
\subsection{Benchmark problem}
\begin{figure}[hbt]
  \centering
   \includegraphics[width=0.9\textwidth]{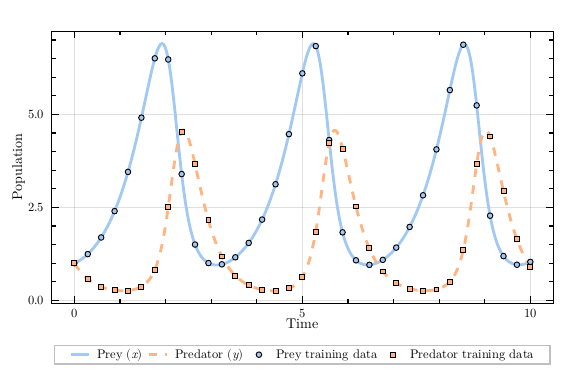}
   \caption{Numerical Lotka-Volterra ODE solution and $N=35$ training data points over time.}
   \label{fig:benchmark_N_lotka_volterra_time}
\end{figure}

\begin{figure}[hbt]
    \centering
   \includegraphics[width=0.9\textwidth]{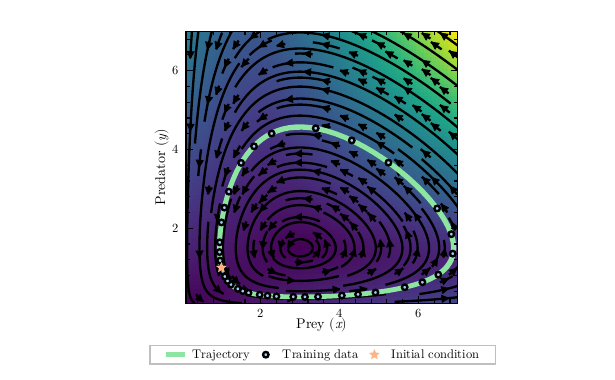}
   \caption{Numerical Lotka-Volterra ODE solution and $N=35$ training data points in the state space}
   \label{fig:benchmark_N_lotka_volterra_state}
\end{figure}

The Lotka-Volterra predator-prey model is a frequently studied problem in data-based ODE learning problems as in the NeuralODE study \citep{chenNeuralOrdinaryDifferential2019a}:
\begin{align}
& \frac{\diff x}{\diff t}=\alpha x-\beta x y \\
& \frac{\diff y}{\diff t}=\gamma x y-\delta y.
\label{eq:Results_LV}
\end{align}
For our benchmark, we replicate the conditions from \citep{chenNeuralOrdinaryDifferential2019a} with \(\alpha=1.5\), \(\beta=1\), \(\gamma=1\), and \(\delta=3\) and initial conditions \(x_0=(1.0, 1.0)^T\).
From a numerical solution of the (forward) ODE problem, we sample $N$ equidistant time points $t_i$ at which we save the two states. These tuples $(t_i, x_i, y_i)$ are then the training data where we assume that we know $(t_0, x_0, y_0)$. Some points and the true solution and the training data over time are shown in \Cref{fig:benchmark_N_lotka_volterra_time}, and the true solution in the state space is shown in \Cref{fig:benchmark_N_lotka_volterra_state}. The training data points are shown as red dots, and the true solution is shown as a black line. The training data points are not equally spaced in the state space, but equally spaced in time. 

We evaluate model performance using three distinct setups to assess different aspects of the learned dynamics. The first setup (\textit{ex-it}) evaluates the in-distribution performance by assessing the model on the original training time interval while using the same initial conditions employed during training. The second setup (\textit{ex-oot}) examines temporal extrapolation capabilities by extending the prediction horizon to twice the training duration while maintaining the same initial conditions. The third setup (\textit{ex-ood}) tests generalization to unseen initial conditions by combining the extended time horizon with different starting points not present in the training data. Table~\ref{tab:performance_metrics} provides an overview of these evaluation strategies. Specifically, we use the following settings:
\begin{center}
    \begin{itemize}
        \item \textit{ex-it}: $t\in (0.0, 7.0)$  and $u_0 = (1.0, 1.0)^T$
        \item \textit{ex-oot}: $t\in (0.0, 14.0)$ and $u_0 = (1.0, 1.0)^T$
        \item \textit{ex-ood}: $t\in (0.0, 14.0)$ and $u_0^{ood} = (0.5, 0.5)^T$.
    \end{itemize}
\end{center}

\subsection{Scenario results}
The scenarios presented below progress from a very general setup to more specialized aspects of ODE learning. Each of the scenarios is evaluated on the three comparison setups, as mentioned before.
\paragraph{S1: Perfect information -- local overfitting}
In this scenario, we investigate how local training data can distort globally accurate approximations when the RHS representation is local. We start by assuming access to the true RHS function throughout the entire state space, which allows us to pretrain the RHS globally optimal coefficients. This global pretraining establishes an initially accurate approximation of the dynamics everywhere.

After establishing these ideally pretrained models, we subsequently fine-tune them using only a single trajectory starting from $x_0=(1.0,1.0)^T$ with dense observations (144 data points per state variable). This setup reveals a critical trade-off: while the models improve their accuracy along the observed trajectory, their performance deteriorates in regions of the state space that are not covered by the training data. This phenomenon demonstrates how local data can "carry away" initially accurate global approximations, creating a bias toward the observed trajectory at the expense of generalization elsewhere.

The results reveal substantial differences among the three modeling approaches. \code achieves high precision with MSE of $1.23 \times 10^{-6}$ for in-training (ex-it) and maintains this accuracy for temporal extrapolation (ex-oot) with $2.83 \times 10^{-6}$. Most notably, it retains strong generalization to unseen initial conditions with an ex-ood MSE of only 0.0119. This superior performance stems from the polynomial chaos basis being ideally matched to the Lotka-Volterra dynamics (quadratic in the states). Since the true RHS consists of polynomial terms ($\alpha x$, $\beta xy$, $\gamma xy$, and $\delta y$), the polynomial basis functions can represent these dynamics exactly. While the ex-ood MSE should theoretically be zero given this exact representation, two factors contribute to the observed error: numerical integration introduces accumulating floating-point errors over the extended time horizon, and the optimization may converge to a near-optimal local minimum that provides an excellent least-squares fit on the training trajectory while deviating slightly from the true global coefficients.

In contrast, NeuralODE and KernelODE exhibit the local overfitting more prominently. Both achieve reasonable training accuracy: KernelODE reaches a MSE of $7.85 \times 10^{-6}$ and NeuralODE achieves a MSE of $2.78 \times 10^{-3}$ on the training trajectory. However, their performance degrades dramatically when tested on new initial conditions. NeuralODE's ex-ood MSE increases to 15.84, while KernelODE's MSE rises to 3.59. These large errors indicate that local fine-tuning has compromised their global approximation capabilities.

The degradation occurs because both models adapt too flexibly to local data. Neural networks use nonlinear activation functions while kernel models employ localized basis functions, both allowing excessive adaptation to the training trajectory. During fine-tuning, gradient updates concentrate on parameters affecting observed data points, while parameters governing unobserved regions drift from their optimal values. Though KernelODE's structured basis functions provide better resilience than NeuralODE, both exhibit substantial generalization errors compared to \code, highlighting that RHS selection is a critical modeling decision.

\begin{figure*}[htbp]
  \centering
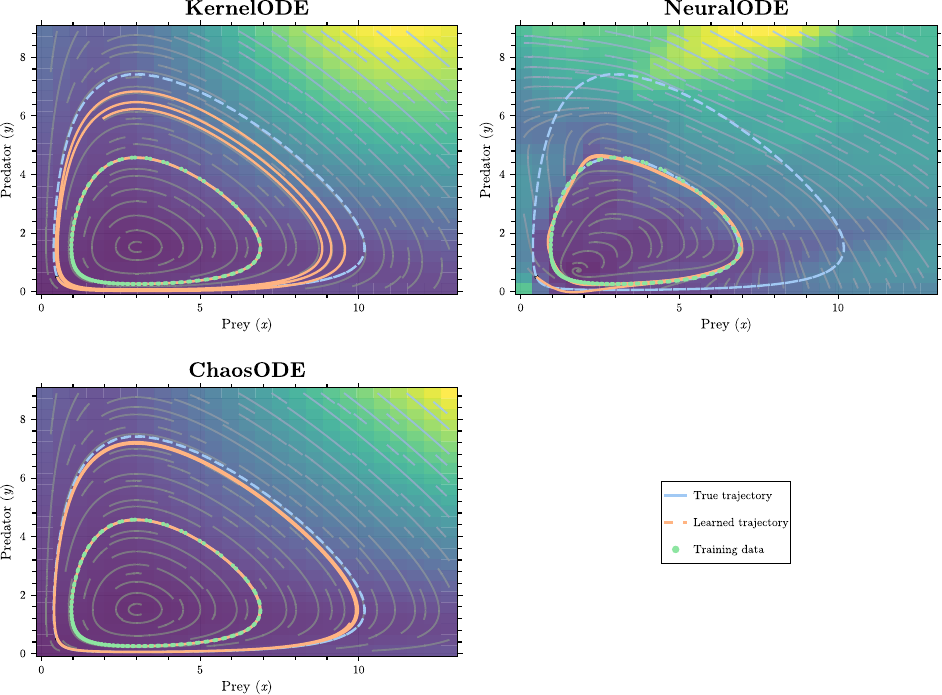
  \caption{Phase space visualization of three different modeling approaches for the Lotka-Volterra system in S1. The heatmap shows vector field magnitude, streamlines represent flow direction, and colored trajectories show model predictions from two initial conditions, resulting in two separate (orange) periodic orbits. Training data points are marked with crosses.}
  \label{fig:S0_state_domain_comparison}
\end{figure*}

Figure~\ref{fig:S0_state_domain_comparison} visualizes these differences in the phase space. The training data appears as green points, with the true trajectories shown in orange and the learned model predictions in orange for two different initial conditions. The NeuralODE solution collapses immediately from the outer periodic orbit into the inner one, producing a trajectory visible only along the training data. KernelODE maintains a more stable extrapolation, though its trajectory slowly spirals inward rather than forming a periodic orbit, deviating moderately from the expected black trajectory. \code produces distinct orbits that closely match the true dynamics, exhibiting only minor inward spiraling. These visualizations directly correspond to the quantitative errors: NeuralODE's immediate divergence reflects its ex-ood MSE of 15.84, KernelODE's gradual drift aligns with its MSE of 3.59, and \code's near-perfect orbits confirm its minimal MSE of 0.0119.

This scenario demonstrates a fundamental challenge in ODE learning: accuracy on training data does not guarantee generalization. Even with perfect initial approximations, local fine-tuning can destroy global accuracy when the chosen basis functions do not match the underlying and required dynamics. Table~\ref{tab:ode_comparison_S1} summarizes these performance metrics, emphasizing that successful extrapolation requires an appropriate model structure.

\begin{table}[htbp]
  \centering
  \ra{0.95}
    \centering
    \begin{tabular}{lccc}
        \toprule
        \textbf{Ansatz} & \textbf{ex-it} & \textbf{ex-oot} & \textbf{ex-ood} \\
        \midrule
        NeuralODE & $2.78 \cdot 10^{-3}$ & $2.77 \cdot 10^{-3}$ & $15.84$ \\
        KernelODE & $7.85 \cdot 10^{-6}$ & $7.93 \cdot 10^{-6}$ & $3.59$ \\
        ChaosODE & $1.23 \cdot 10^{-6}$ & $2.83 \cdot 10^{-6}$ & $0.0119$ \\
        \bottomrule
    \end{tabular}
    \caption{Performance comparison in S1. In-training mean squared error (ex-it), out-of-training mean squared error (ex-oot), and out-of-distribution mean squared error (ex-ood) for different ODE models.}
    \label{tab:ode_comparison_S1}
\end{table}
\FloatBarrier

\paragraph{S2: Data effect -- more is not better?}
In this scenario, we examine how effectively the various RHS choices can learn the generating dynamics of the Lotka-Volterra system. We increased the number of training data points from 10 to 500. For each training point size, we run 10 independent optimizations with different random initializations to account for starting dependent training outcomes.


\begin{figure}[hbtp]
    \centering
    \includegraphics[width=0.95\linewidth]{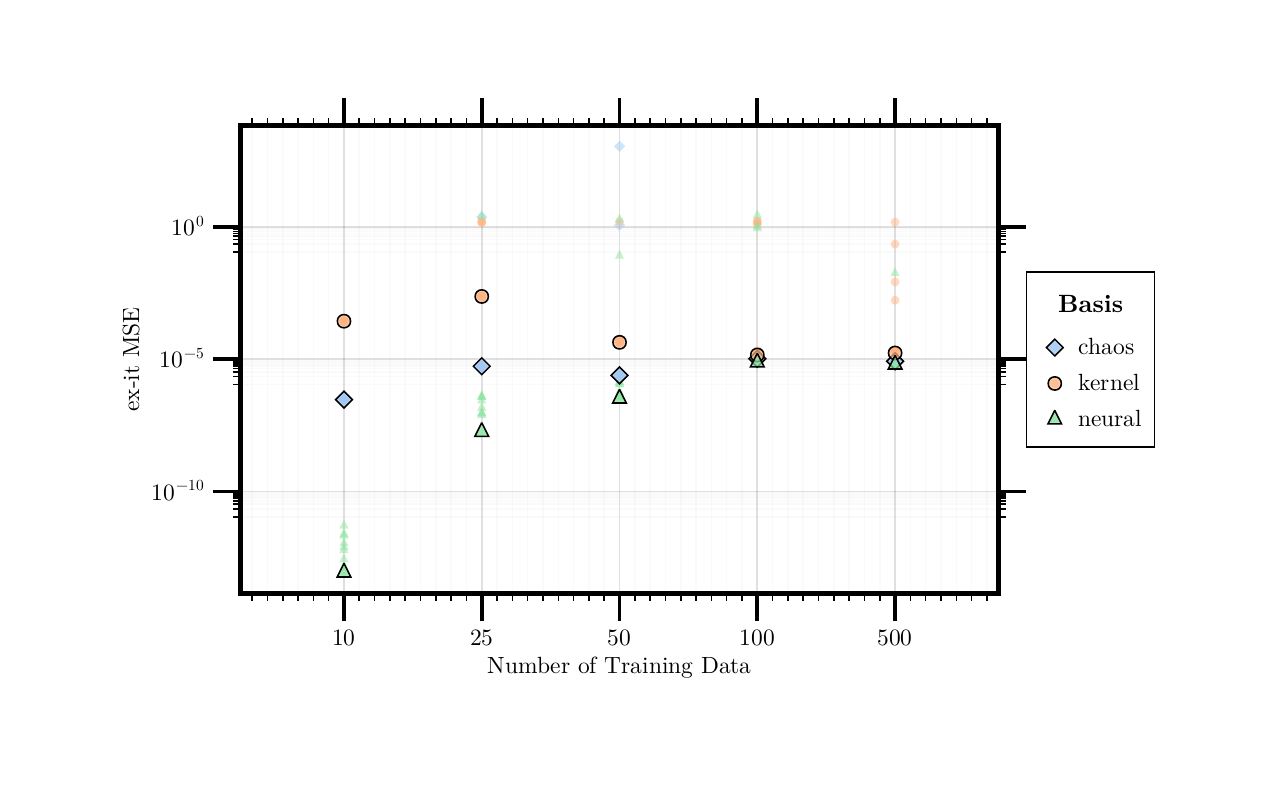}
    \caption{Convergence of the MSE \textit{ex-it} error as training data size increases from 10 to 100 samples. Training data contains no noise ($\sigma=0.0$). Stars indicate the best-performing model for each combination of training data size and basis function.}
    \label{fig:S2_data_convergence_ex-it_0.0}
\end{figure}

First, we analyze the behavior in the \textit{ex-it} setup, where we test on the training data points. The observed error represents the training error and demonstrates how well the model can fit the given data points without requiring generalization. In \Cref{fig:S2_data_convergence_ex-it_0.0}, we observe that NeuralODE particularly benefits from its locality and high flexibility. The model can accurately fit individual data points, even with limited training data. The other two methods do not show significant improvement as the data volume increases. However, with insufficient data points, we may violate the Nyquist sampling theorem \citep{nyquist1928} for oscillatory behavior, potentially yielding solutions that pass through data points at multiples of the true frequency.

Generally, we would expect that methods with stronger structural assumptions require fewer data points to achieve good generalization performance. We do see that more data points lead to less spread among all three RHS basis choices. The low flexibility of KernelODE and its mismatch to Lotka–Volterra dynamics make it the weakest in \textit{ex-it}.
%
  \begin{figure}[hbtp]
    \centering
    \includegraphics[width=0.95\linewidth]{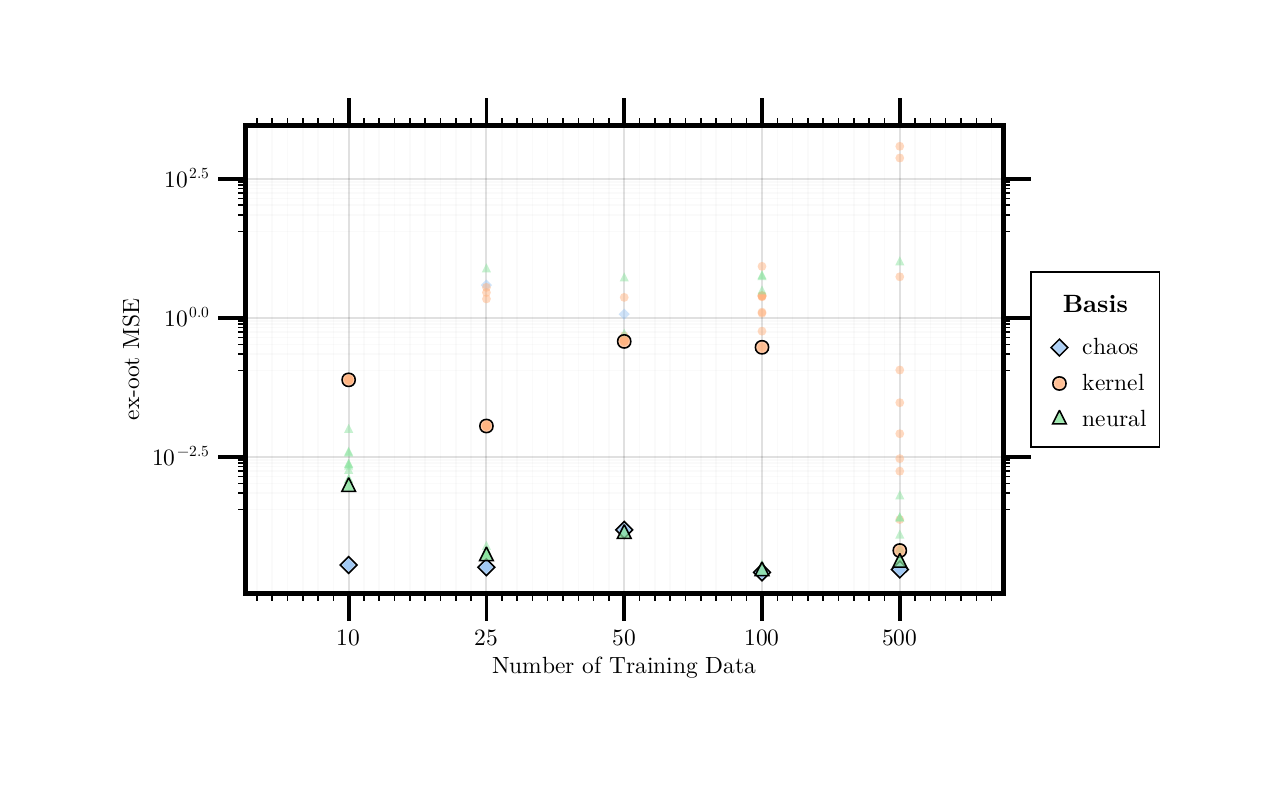}
    \caption{Convergence of the MSE \textit{ex-oot} error as training data size increases from 10 to 100 samples. Training data contains no noise ($\sigma=0.0$). Stars indicate the best-performing model for each combination of training data size and basis function.}
    \label{fig:S2_data_convergence_ex_oot_0.0}
  \end{figure}
\smallskip

For tests in the \textit{ex-oot} setup (see \cref{fig:S2_data_convergence_ex_oot_0.0}), we first observe that we are no longer within the training dataset, and the average MSE range becomes higher by several magnitudes. However, as the ChaosODE RHS is ideally able to represent the true dynamics, it succeeds in learning the most accurate representation within our comparison. In particular, a higher data density improves NeuralODE \textit{ex-oot} performance, aligning with established scaling laws \citep{kaplan2020scaling}. In these experiments, the best KernelODE solution performs worse than the best NeuralODE solution for all numbers of training data.
\smallskip

 \begin{figure}[hbtp]
    \centering
    \includegraphics[width=0.95\linewidth]{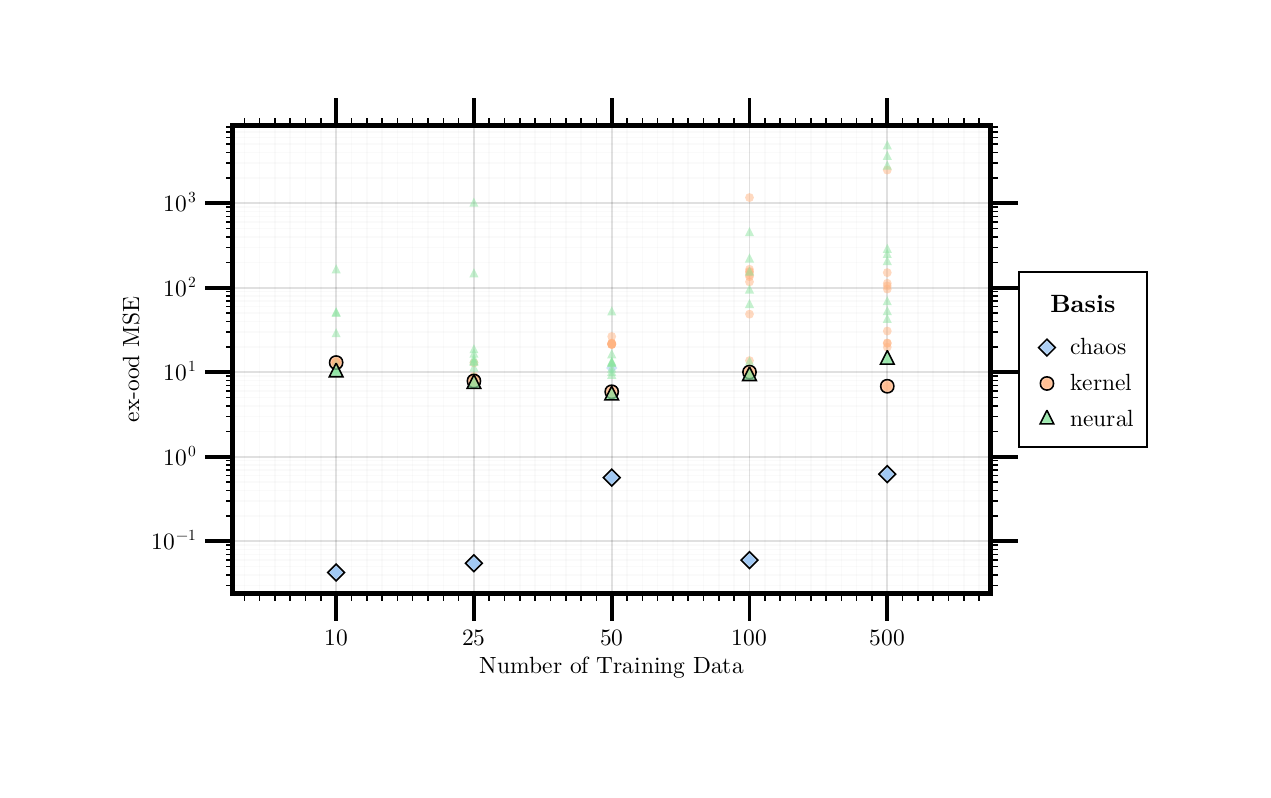}
    \caption{MSE \textit{ex-ood} error as a function of training data size. Training data contains no added noise ($\sigma=0.0$). Stars indicate the best-performing model for each combination of training data size and basis function.}   
    \label{fig:S2_data_convergence_ex_ood_0.0}
  \end{figure}

Interpolation in dense data regimes is well established and can be achieved using many machine learning tools. However, generalization remains one of the biggest challenges \citep{rohlfs2025generalization}. In the \textit{ex-ood} setup, the model is evaluated on trajectories from new initial conditions not seen during training. Without additional assumptions, no model has theoretical guarantees for generalizing to these cases, particularly when predicting system behavior from initial conditions requires extrapolation beyond the training data range.

The assumptions that explicitly or implicitly inform the modeling approach differ across methods: for NeuralODE, these include all hyperparameter choices such as activation functions and the number of layers; for KernelODE, they primarily involve the kernel choice and kernel hyperparameters; for \code, our key assumption is the highest polynomial degree. The three basis function examples clearly demonstrate that different basis choices lead to different extrapolation properties. In this noise-free scenario, we therefore demonstrate that choosing a basis function well-suited to approximate the observed dynamics enables successful extrapolation beyond the training domain.

As shown in \Cref{fig:S2_data_convergence_ex_ood_0.0} for the \textit{ex-ood} case, \code solutions achieve the best extrapolation performance across all training data sizes. Since the training data contains no noise, we attribute the increased error with larger datasets to the transition from multiple-shooting to single-shooting optimization. With more data points, the multiple-shooting approach introduces higher continuity errors between segments that are not adequately resolved during the final optimization step. In general, we observe that selecting an appropriate basis function at the beginning of the learning process --- in our case, a polynomial basis—is essential for extrapolation that goes beyond random guessing.

\begin{figure}[hbtp]
  \centering
  \includegraphics[width=0.95\linewidth]{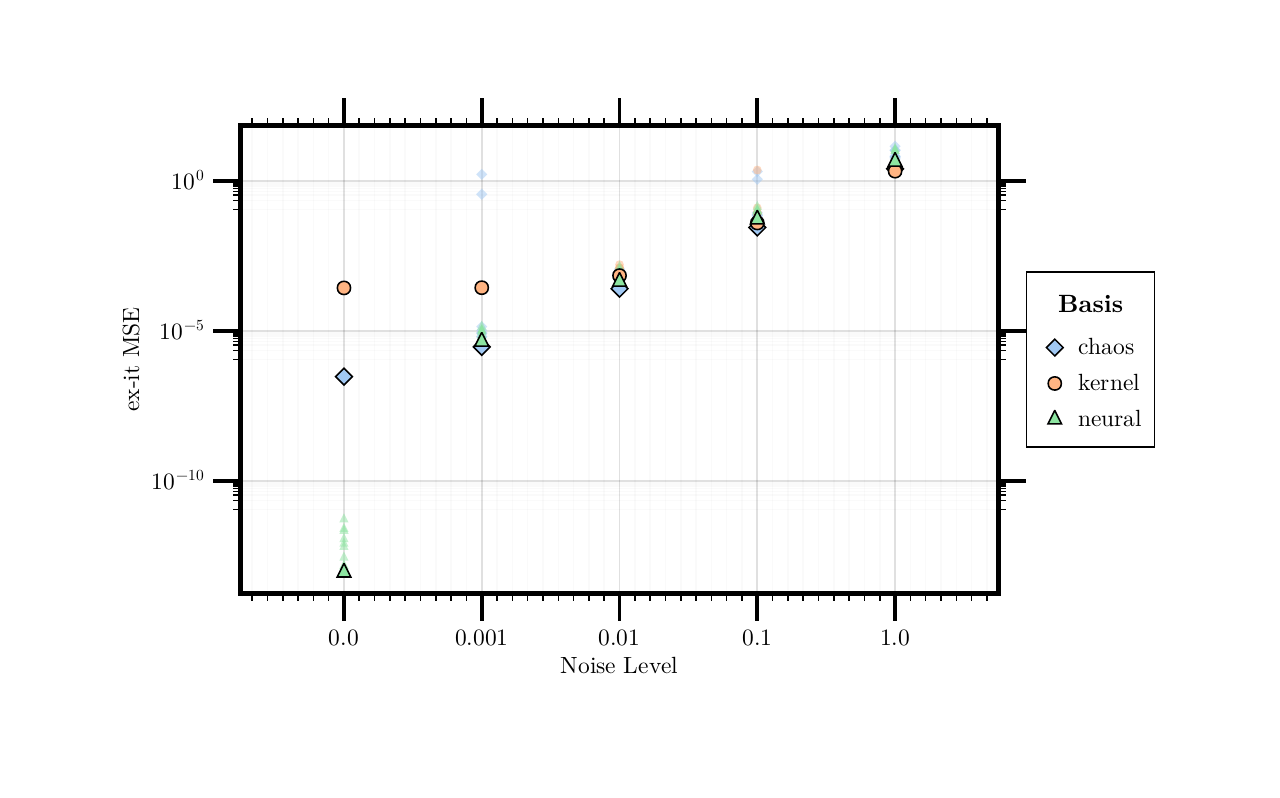}
  \caption{Comparison of MSE on training data (\textit{ex-it}) across different noise levels and methods, trained on 10 data points.}
  \label{fig:ex_it_10}
\end{figure}

\begin{figure}[hbtp]
  \centering
  \includegraphics[width=0.95\linewidth]{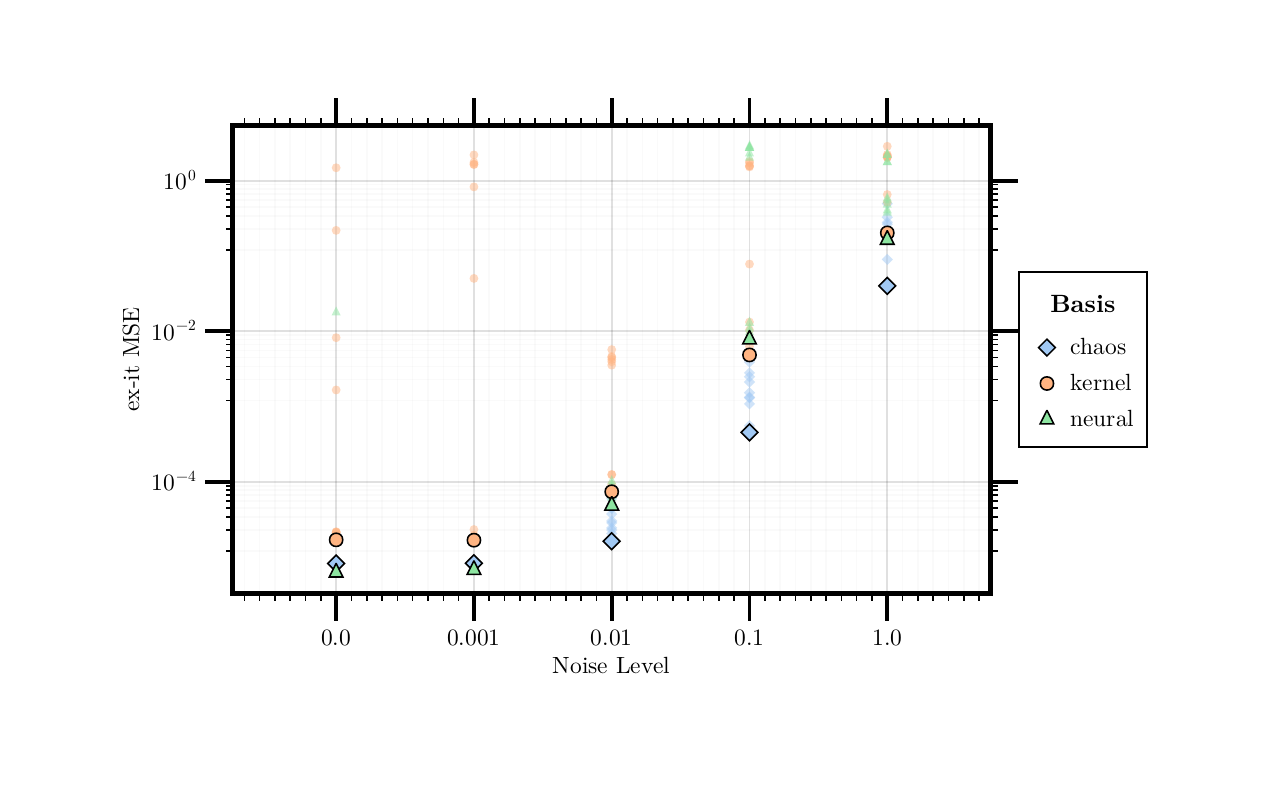}
  \caption{Comparison of MSE on training data (\textit{ex-it}) across different noise levels and methods, trained on 500 data points.}
  \label{fig:ex_it_500}
\end{figure}

\begin{figure}[hbtp]
    \centering
    \includegraphics[width=0.95\linewidth]{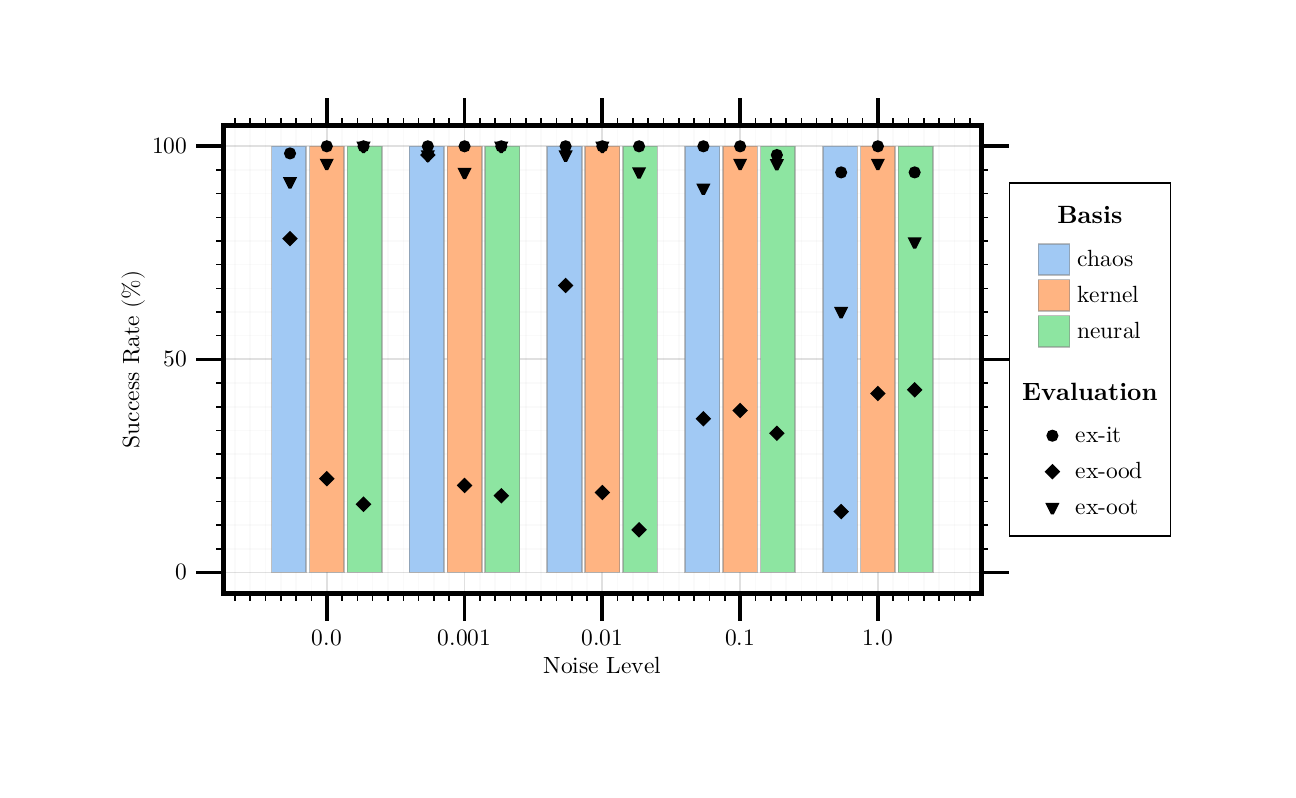}
    \caption{Success rates (percentage of solutions with MSE $< 10.0$) across different noise levels and training data sizes.}
    \label{fig:S3_stability10}
\end{figure}

\paragraph{S3: Noise makes training more difficult.}
Measured data typically contains noise, including measurement errors represented by the random component $\varepsilon$. In the S3 experiments, we model this noise as $\varepsilon \sim \mathcal{N}(0,\sigma)$ with varying noise levels $\sigma$. \Cref{fig:ex_it_10,fig:ex_it_500} show training point errors across different noise levels for datasets with 10 and 500 data points, respectively. As noise increases, the minimum achieved MSE rises across all methods, as to be expected. The 500-data-point case reveals increased variance, particularly for KernelODE, though no divergent values or optimization failures occur within the displayed range.
To assess optimization robustness comprehensively in S3, we analyze success rates across different conditions. We define training failure as achieving MSE $> 10.0$, evaluating all combinations of training point quantities and ten random seeds (providing different initial values and noise realizations). \Cref{fig:S3_stability10} presents these results.
The \textit{ex-it} (in-training time) performance remains consistently high, with success rates between 95\% and 100\% across all conditions. This demonstrates robust training stability despite noise variations, different realizations, and varying data quantities. The \textit{ex-oot} (out-of-training time) case naturally yields higher MSEs, resulting in lower success rates.
\code achieves the highest success rates, particularly at low noise levels ($\sigma = 0.0, 0.001, 0.01$). This superior performance results from the perfect fit of the basis for the underlying polynomial dynamics. However, high noise ($\sigma = 1.0$) causes a notable drop in \code success rates.
Noise creates competing effects on the training dynamics. The MSE loss function should theoretically center solutions within the empirical distribution by accounting for symmetric noise. However, increased noise generates multiple parameter configurations with identical loss values. This condition makes the optimization problem less well-posed and can cause oscillations between training points, especially as we have not fine-tuned the methods to the higher noise, \eg tuning the regularization, bandwidth for KernelODE or optimizer metaparameters. In contrast, NeuralODE and KernelODE show slight stability improvements with additional noise. This improvement potentially results from regularization effects that are not specifically optimized for the noise levels or data quantities tested. The results for the \textit{ex-oot} and \textit{ex-ood} are provided in the appendix \cref{secA1}.

\FloatBarrier
\paragraph{S4: Orthonormal vs. Non-orthonormal Basis}

Section S4 examines how basis construction affects \code performance in all evaluation metrics. We compare results using a data-dependent orthonormal basis against a non-orthonormal polynomial basis (such as the Vandermonde basis) under identical training conditions. Again, 10 runs were performed for each configuration.

The orthonormal basis creates a well-conditioned optimization problem by ensuring that the basis functions have unit norm and are mutually orthogonal. This conditioning avoids the numerical difficulties associated with the Vandermonde basis, where the basis functions can have vastly different scales and high correlation (\cite{GAUTSCHI1983293}). The resulting system behaves similarly to what would be achieved by preconditioning the non-orthonormal case, but achieves this conditioning directly through basis construction. Similar results were reported in other regression-like scenarios (\cite{li2011orthogonal}).

\begin{figure}[hbtp]
    \centering
    \includegraphics[width=0.95\linewidth]{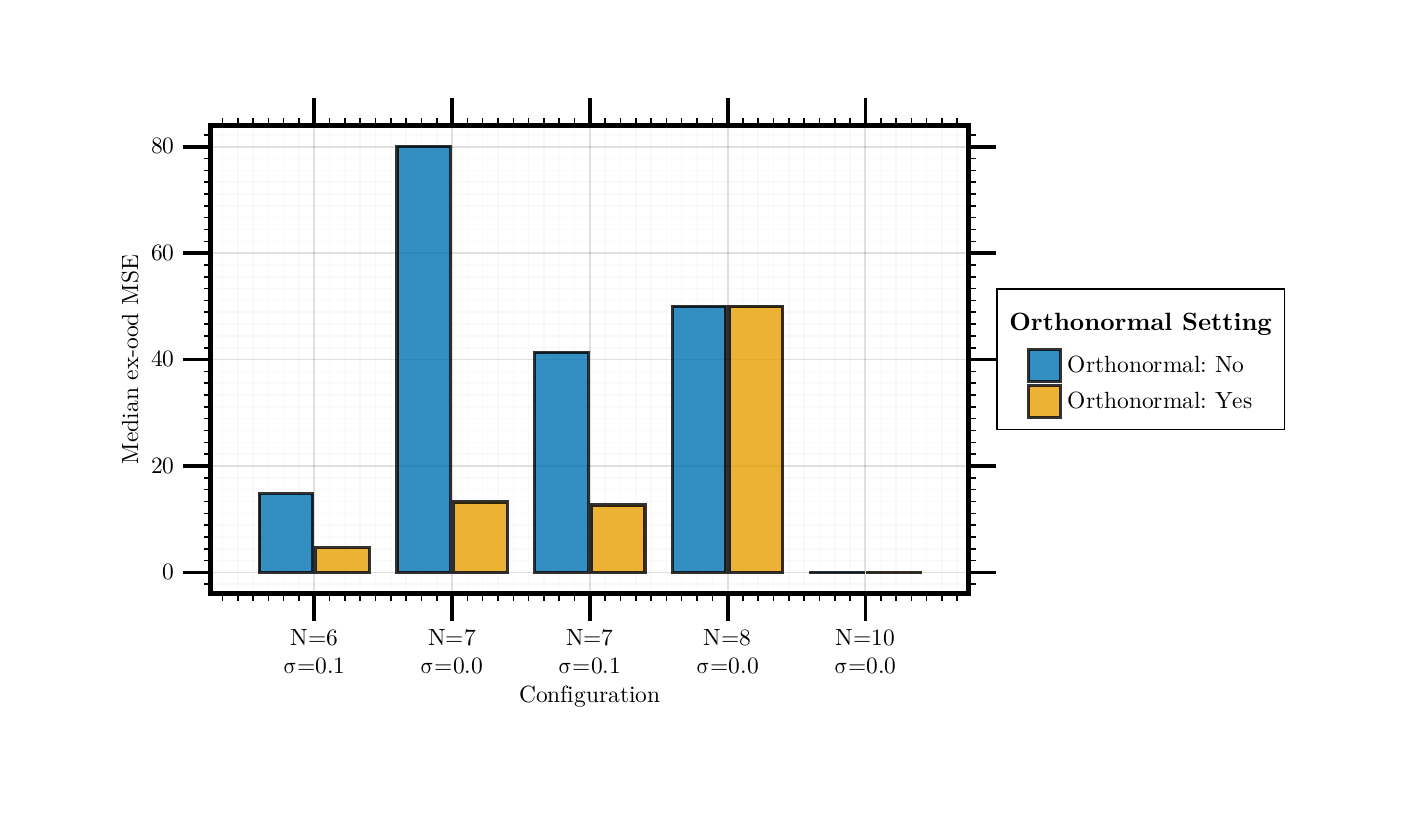}
    \caption{Out-of-distribution performance (\textit{ex-ood}) comparing orthonormal versus non-orthonormal basis construction across noise and data conditions. The orthonormal basis shows consistent improvements, particularly in low-data regimes.}
    \label{fig:S4_orthonormality}
\end{figure}

\begin{figure}[hbtp]
    \centering
    \includegraphics[width=0.95\linewidth]{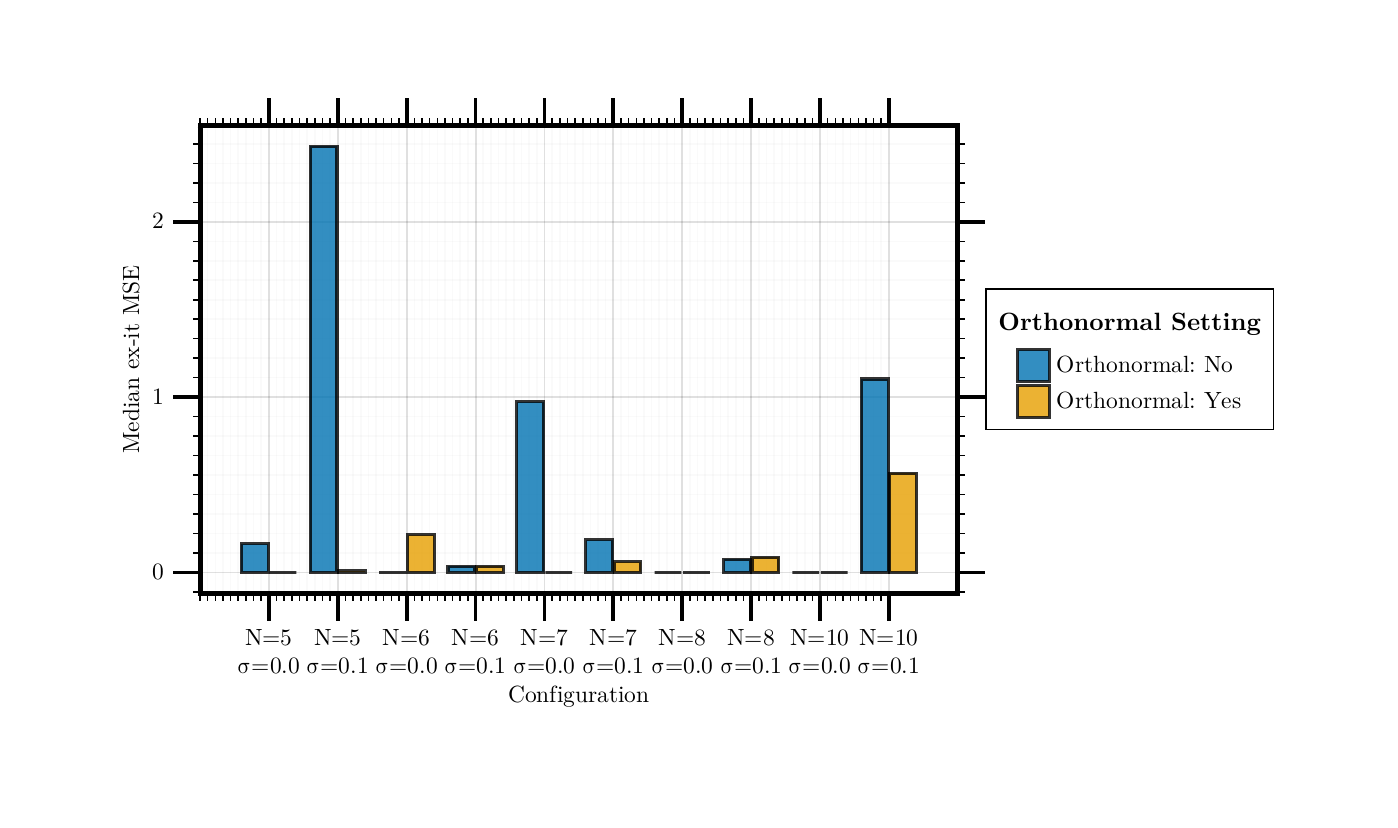}
    \caption{Training performance (\textit{ex-it}) comparing orthonormal versus non-orthonormal basis construction across different noise levels and data quantities. Orthonormal basis construction reduces training error, with largest improvements observed in low-data conditions.}
    \label{fig:S4_exit}
\end{figure}

\begin{figure}[hbtp]
    \centering
    \includegraphics[width=0.95\linewidth]{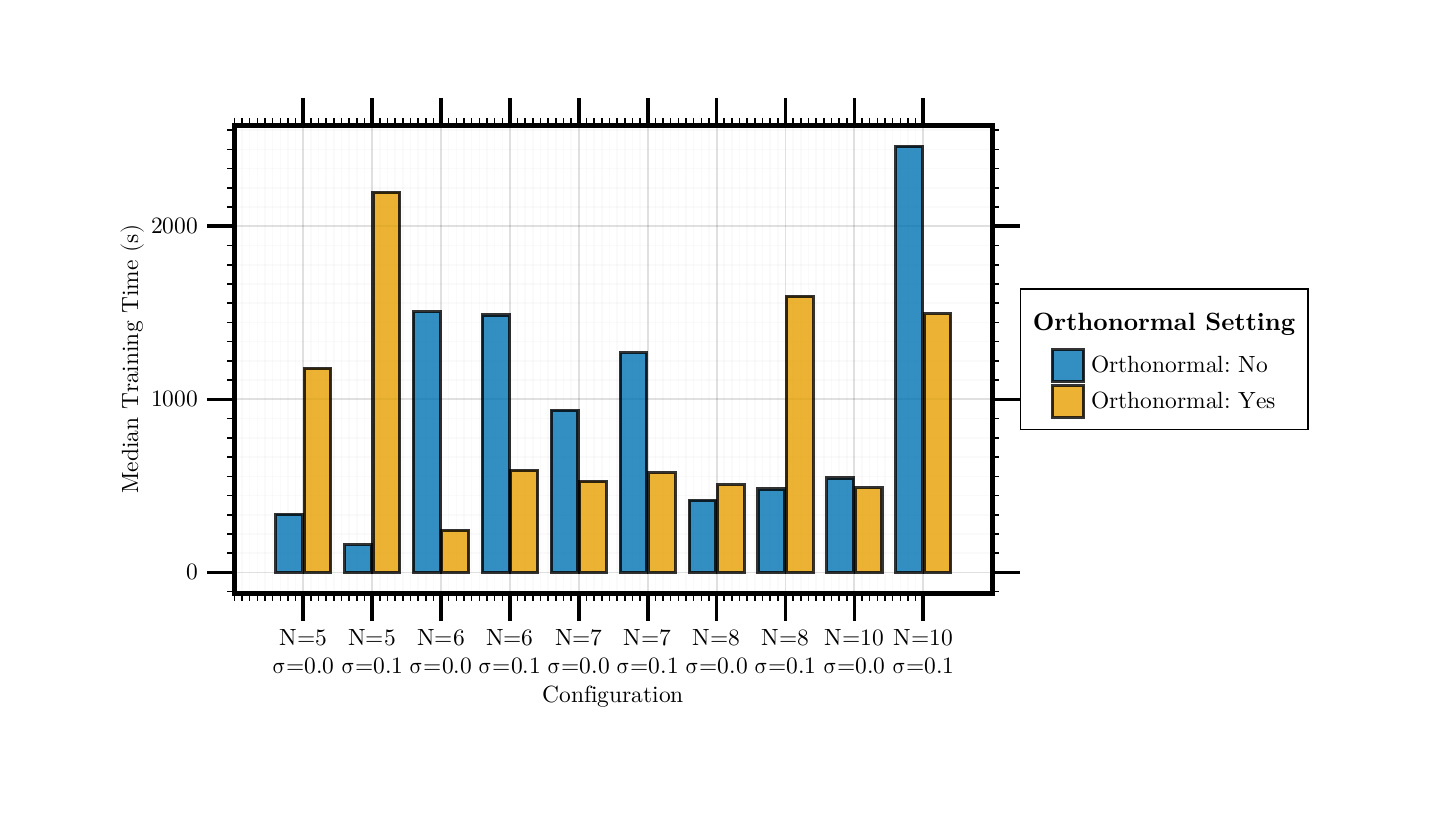}
    \caption{Median training time comparing orthonormal versus non-orthonormal basis construction across different noise levels and data quantities. Orthonormal basis construction often reduces computational time in the low data regime.}
    \label{fig:S4_timing}
\end{figure}

The orthonormal basis consistently reduces errors across all evaluation scenarios. \Cref{fig:S4_orthonormality} shows substantial improvements in out-of-distribution performance, particularly in low-data regimes. Training performance exhibits similar patterns (\Cref{fig:S4_exit}), with the orthonormal basis achieving lower MSE values in noise and data conditions. These improvements come with reduced computational cost; \Cref{fig:S4_timing} demonstrates faster convergence times, especially in data-limited scenarios.

\begin{figure}[hbtp]
    \centering
    \includegraphics[width=0.95\linewidth]{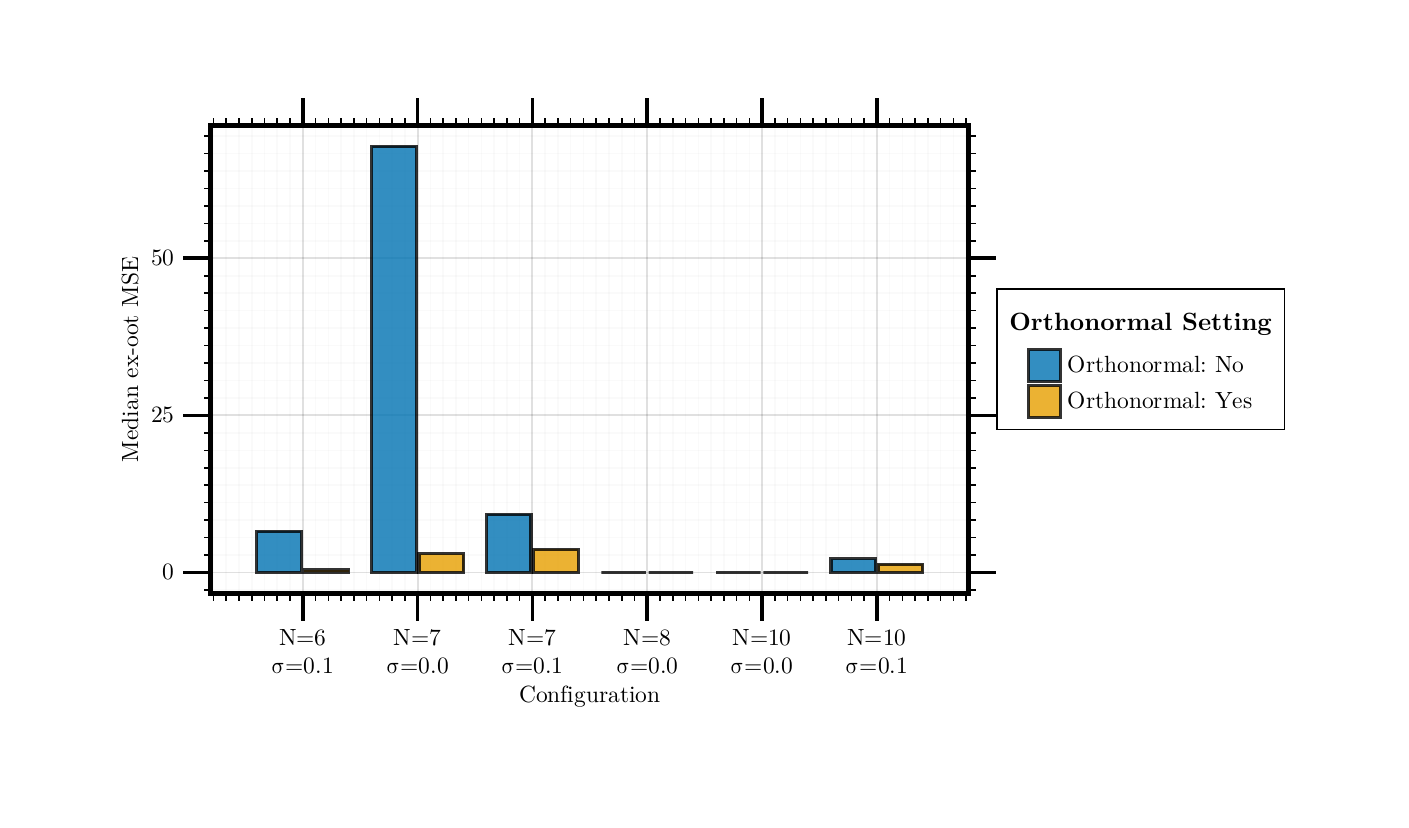}
    \caption{Training performance (\textit{ex-oot}) comparing orthonormal versus non-orthonormal basis construction across different noise levels and data quantities. Orthonormal basis construction reduces training error, with largest improvements observed in low-data conditions.}
    \label{fig:S4_ex-oot}
\end{figure}

\FloatBarrier

\FloatBarrier

\section{Conclusion}
\label{sec:conclusion}
We introduce the novel ChaosODE (\code) approach, an orthonormal polynomial chaos-based approximation method to learn arbitrary ODE dynamics from time-series state observations. We demonstrated that selecting problem-tailored ansatz spaces for the right-hand side (RHS) provides substantial benefits in training robustness across varying noise levels and data availability. Through a systematic comparison with two established approaches, NeuralODE and KernelODE, we addressed three foundational questions about structure-aware universal differential equation (UDE) learning:
\begin{itemize}
   \item \textbf{RQ1 (Scarce and noisy data):} How do current UDE learning techniques perform when training data is both scarce and noisy? We found \code's superior performance under challenging data conditions, showing robust learning capabilities even with limited and noisy observations.
   \item \textbf{RQ2 (Extrapolation and generalization):} What extrapolation and generalization capabilities do KernelODE and NeuralODE exhibit, and how does \code's global approximation structure compare? We showed that \code's global orthonormal polynomial approximation structure enables effective extrapolation to unseen initial state values, providing clear advantages over local approximation methods.
   \item \textbf{RQ3 (Orthonormal structure benefits):} Does the orthonormal structure of the dynamics representation facilitate learning in highly non-linear systems? We developed a robust optimization pipeline that addresses numerical instability issues in ODE learning and demonstrated that the orthonormal basis yields more effective learning compared to a non-orthonormal polynomial basis.
\end{itemize}
This research challenges the common practice of using NeuralODE as a default for autonomous continuous-time UDE learning, advocating instead for architectural decisions informed by problem structure. We highlight that the selection of the RHS basis should align with the system dynamics, as it implicitly introduces inductive bias. Our results indicate that carefully choosing the approximation space improves learning efficiency and model performance in ODE tasks.

\FloatBarrier


\subsubsection*{Acknowledgments}
We thank the Deutsche Forschungsgemeinschaft (DFG, German Research Foundation) for supporting this work by funding -- EXC2075 -- 390740016 under Germany's Excellence Strategy and the Collaborative Research Centre SFB 1313, Project Number 327154368. We acknowledge the support by the Stuttgart Center for Simulation Science (SimTech).

\subsubsection*{Data Availability Statement}
Code (in julia) to reproduce all figures and results will be publicly available upon acceptance at \url{https://github.com/NilsWildt/CODE-Global-ODE-learning}. All findings can be reproduced by running the provided scripts as described in the repository documentation.
\bibliography{bibliography}
\bibliographystyle{tmlr}

\clearpage
\appendix

\section{Supplementary material S3: Further analysis of the noise cases}\label{secA1}
\begin{figure}[hbtp]
  \centering
  \includegraphics[width=0.95\linewidth]{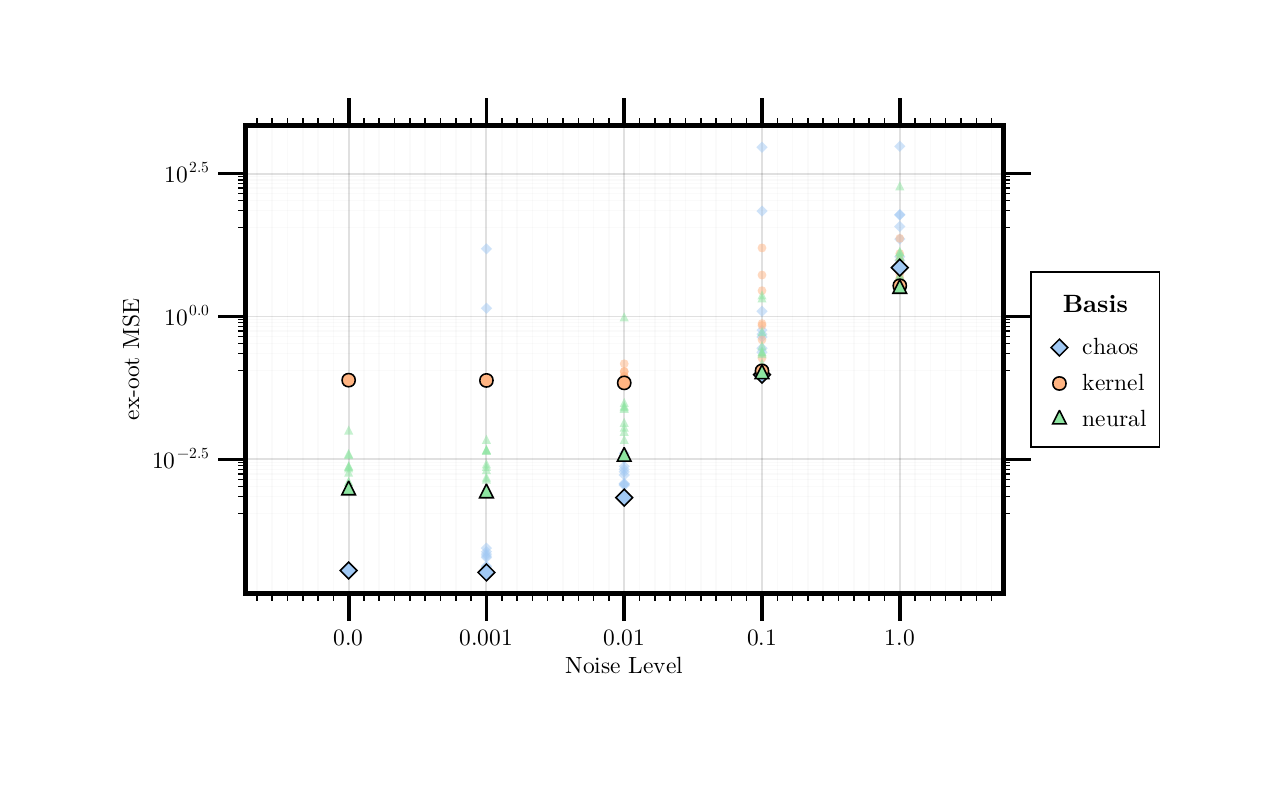}
  \caption{Comparison of extrapolation MSE (\textit{ex-oot}) across different noise levels and methods, trained on 10 data points.}
  \label{fig:ex_oot_10}
\end{figure}

\begin{figure}[hbtp]
  \centering
  \includegraphics[width=0.95\linewidth]{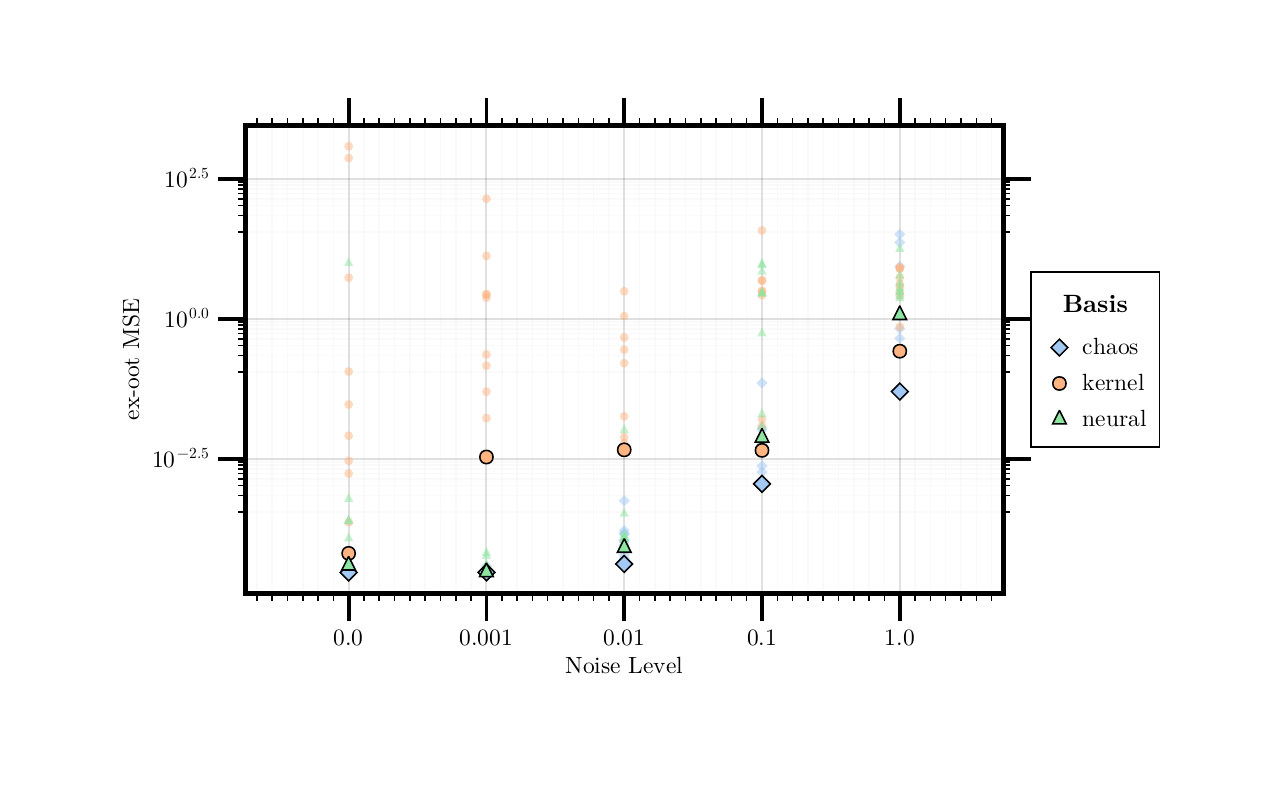}
  \caption{Comparison of extrapolation MSE (\textit{ex-oot}) across different noise levels and methods, trained on 500 data points.}
  \label{fig:ex_oot_500}
\end{figure}

\begin{figure}[hbtp]
  \centering
  \includegraphics[width=0.95\linewidth]{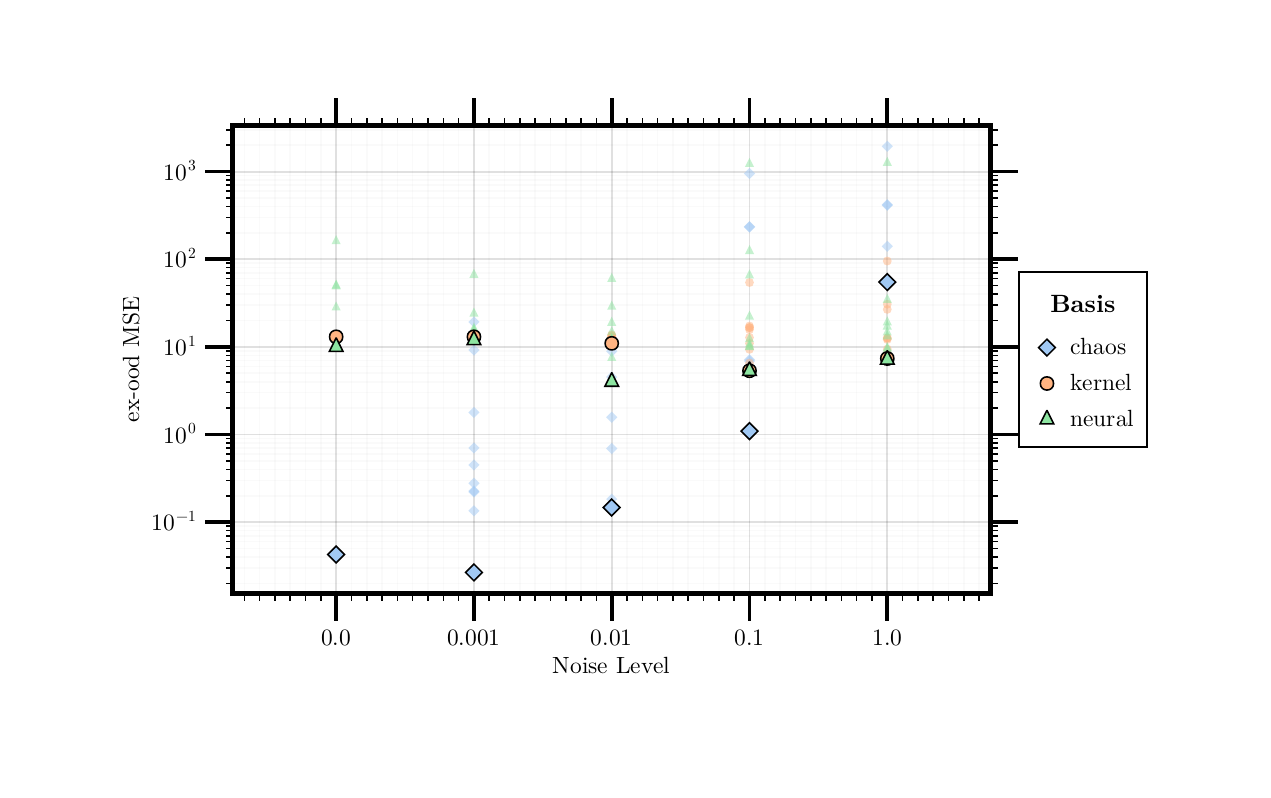}
  \caption{Comparison of out-of-distribution MSE (\textit{ex-ood}) across different noise levels and methods, trained on 10 data points.}
  \label{fig:ex_ood_10}
\end{figure}

\begin{figure}[hbtp]
  \centering
  \includegraphics[width=0.95\linewidth]{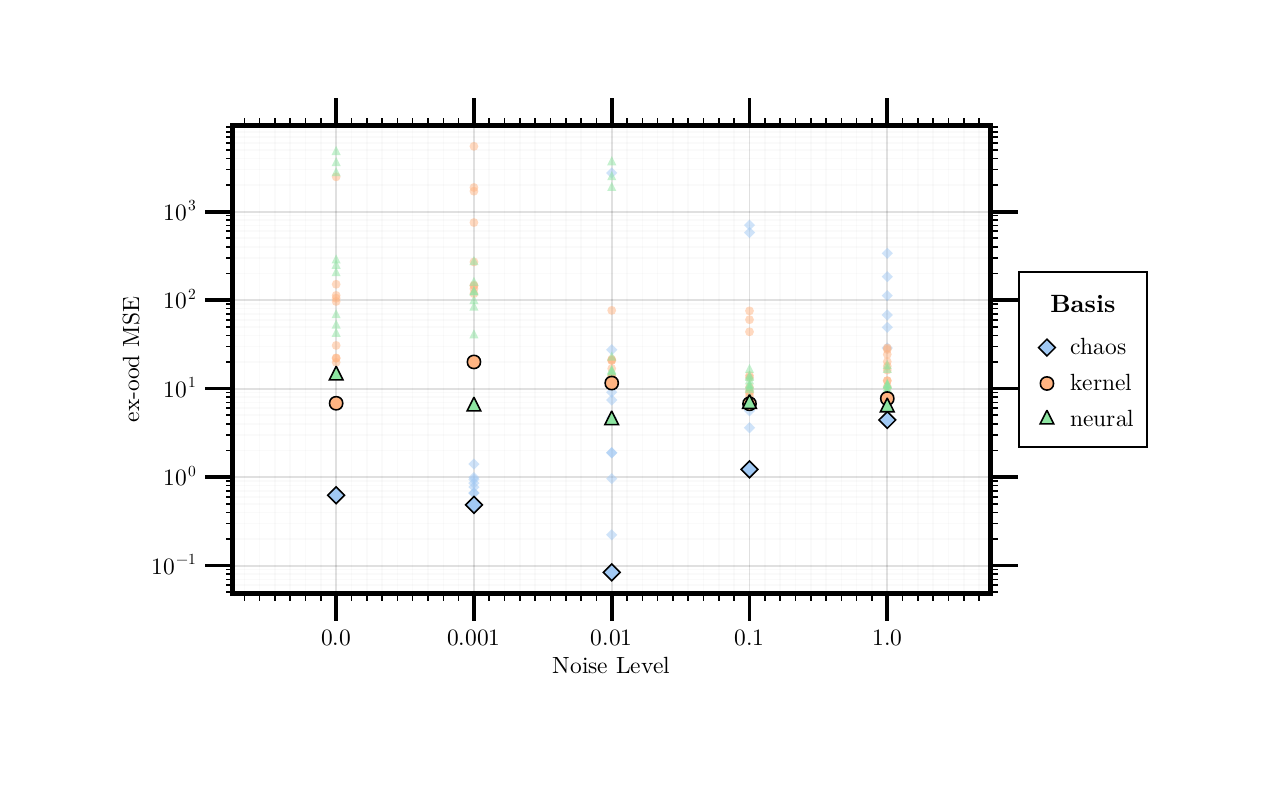}
  \caption{Comparison of out-of-distribution MSE (\textit{ex-ood}) across different noise levels and methods, trained on 500 data points.}
  \label{fig:ex_ood_500}
\end{figure}




\FloatBarrier

\section{Used Packages}\label{secA2}
For the implementation we rely on Julia \citep{Julia-2017} using the following packages (among others):
\begin{itemize}
\item AlgebraOfGraphics.jl for grammar-based data visualization
\item ComponentArrays.jl for structured array handling \citep{ComponentArrays2021}
\item DifferentiationInterface.jl as an interface to automatic differentiation backends \citep{Dalle_Hill_DifferentiationInterface_2025}
\item DataFrames.jl and DataFramesMeta.jl for tabular data handling \citep{BouchetValatKaminski2023}
\item DifferentialEquations.jl ecosystem (OrdinaryDiffEq, OrdinaryDiffEqTsit5, SciMLBase) for solving differential equations \citep{Rackauckas2017}
\item Distributions.jl for probability distributions \citep{Besancon2021}
\item DrWatson.jl for scientific project management \citep{Datseris2020}
\item ForwardDiff.jl for forward-mode automatic differentiation \citep{Revels2016}
\item JLD2.jl for data serialization \citep{JLD2}
\item Lux.jl for neural network implementations \citep{Pal2023}
\item Makie.jl and CairoMakie.jl for plotting and visualization \citep{DanischKrumbiegel2021}
\item Optimization.jl for optimization algorithms \citep{OptimizationJL2023}
\item Zygote.jl for automatic differentiation \citep{Innes2018}
\end{itemize}

\FloatBarrier

\end{document}